\documentclass[journal]{IEEEtran}
\ifCLASSINFOpdf
\usepackage{graphicx}
\usepackage{caption}
\usepackage{stfloats}
\usepackage{float}
\usepackage{amsmath}
\usepackage{amssymb}
\usepackage{url}
\usepackage{subfigure}
\usepackage{hhline}
\usepackage{times}
\usepackage{epsfig}
\usepackage{multirow}
\usepackage{array}
\usepackage{color}
\usepackage{booktabs}
\usepackage{hyperref}
\hypersetup{breaklinks=true,colorlinks}
\usepackage[noadjust]{cite}
\newcommand{\etal}{{\emph{et al.}}}

\begin{document}
\title{Unsupervised Low-Light Image Enhancement\\
via Histogram Equalization Prior}
\author{Feng~Zhang, Yuanjie~Shao, Yishi~Sun, Kai~Zhu, Changxin~Gao,and~Nong~Sang\\

\thanks{Feng~Zhang,
        Yuanjie~Shao,
        Yishi~Sun,
        Kai~Zhu,
        Changxin~Gao,
        and~Nong~Sang are with the National Key Laboratory of Science and Technology on Multispectral Information Processing, School of Artificial Intelligence and Automation, Huazhong University of Science and Technology, Wuhan 430074, China (e-mail: fengzhangaia@hust.edu.cn; cgao@hust.edu.cn; nsang@hust.edu.cn).}
\thanks{Correspondence addressed to: Nong Sang (nsang@hust.edu.cn)}
\thanks{Manuscript received April 9, 2021; revised August 26, 2021.}}

\markboth{Journal of \LaTeX\ Class Files,~Vol.~14, No.~8, August~2022}%
{Shell \MakeLowercase{\textit{et al.}}: Bare Demo of IEEEtran.cls for IEEE Journals}

\maketitle
\begin{abstract}
Deep learning-based methods for low-light image enhancement typically require enormous paired training data, which are impractical to capture in real-world scenarios. Recently, unsupervised approaches have been explored to eliminate the reliance on paired training data. However, they perform erratically in diverse real-world scenarios due to the absence of priors. To address this issue, we propose an unsupervised low-light image enhancement method based on an effective prior termed histogram equalization prior (HEP). Our work is inspired by the interesting observation that the feature maps of histogram equalization enhanced image and the ground truth are similar. Specifically, we formulate the HEP to provide abundant texture and luminance information. Embedded into a Light Up Module (LUM), it helps to decompose the low-light images into illumination and reflectance maps, and the reflectance maps can be regarded as restored images. However, the derivation based on Retinex theory reveals that the reflectance maps are contaminated by noise. We introduce a Noise Disentanglement Module (NDM) to disentangle the noise and content in the reflectance maps with the reliable aid of unpaired clean images. Guided by the histogram equalization prior and noise disentanglement, our method can recover finer details and is more capable to suppress noise in real-world low-light scenarios. Extensive experiments demonstrate that our method performs favorably against the state-of-the-art unsupervised low-light enhancement algorithms and even matches the state-of-the-art supervised algorithms. The code is available at \url{https://github.com/fengzhang427/HEP}.
\end{abstract}


%
\IEEEpeerreviewmaketitle

\section{Introduction}
\IEEEPARstart{I}{mages} captured under low-light conditions often suffer from poor visibility, unexpected noise, and color distortion. 
In order to take high-quality images in low-light conditions, several operations including setting long exposures, high ISO, and flash are commonly applied. However, solely turning up the brightness of dark regions will inevitably amplify image degradation.
To further mitigate the degradation caused by low-light conditions, several traditional methods have been proposed. Histogram Equalization (HE)~\cite{pizer1990contrast} rearranges the pixels of the low-light image to improve the dynamic range of the image. Retinex-based methods~\cite{wang2013naturalness, wang2014variational} decompose the low-light images into illumination and reflection maps and obtain the intensified image by fusing the enhanced reflection map and illumination map. Dehazing-based methods~\cite{dong2011fast, li2015low} regard the inverted low-light image as a haze image and improve visibility by applying dehazing. Although these methods can improve brightness, especially for dark pixels, they barely consider realistic lighting factors, often making the enhanced results visually tenuous and inconsistent with the actual scene.

\begin{figure}
    \begin{center}
    \subfigure[Input]{ 
        \includegraphics[width=0.48\linewidth]{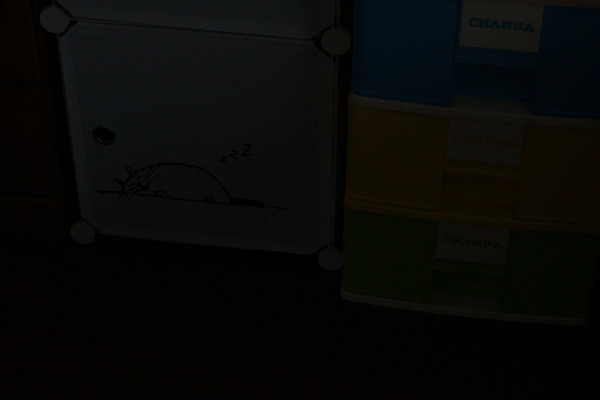}
    }\hspace*{-2mm}
    \subfigure[Zero-DCE~\cite{guo2020zero}]{
        \includegraphics[width=0.48\linewidth]{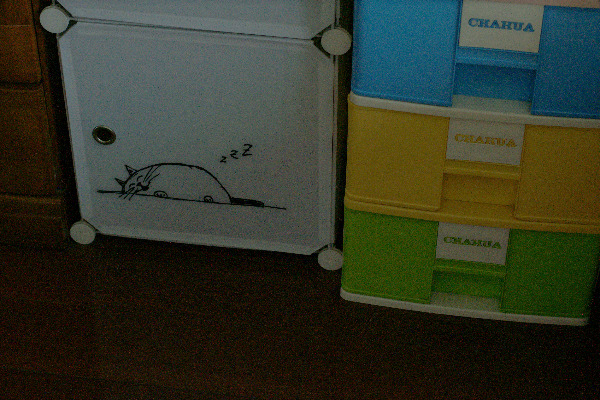}
    }
    \subfigure[EnglightenGAN~\cite{jiang2021enlightengan}]{
        \includegraphics[width=0.48\linewidth]{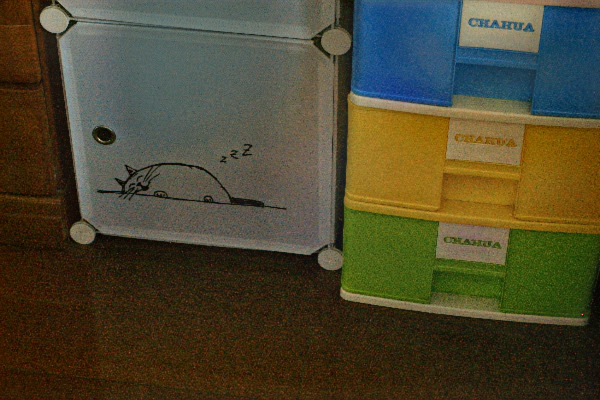}
    }\hspace*{-2mm}
    \subfigure[Ours]{
        \includegraphics[width=0.48\linewidth]{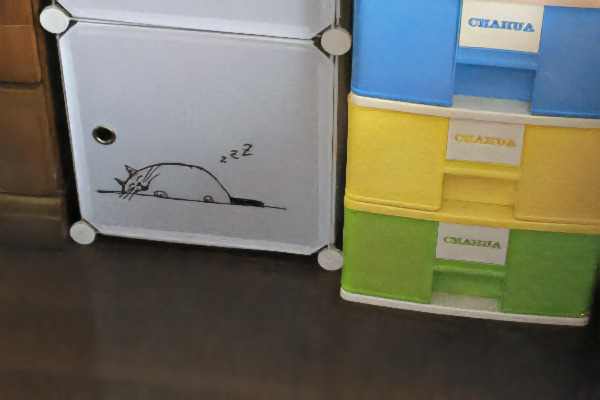}
    }
    \caption{Visual results of the proposed method compared with the state-of-the-art unsupervised low-light enhancement methods. The low-light image of (a) is from EnlightenGAN test set~\cite{jiang2021enlightengan}.}
    \label{fig:intro}
    \end{center}
\end{figure}

Recently, Deep Convolutional Neural Networks (CNNs) set the state-of-the-art in low-light image enhancement. Compared with traditional methods, the CNNs learn better feature representation to obtain enhanced results with superior visual quality, which benefits from the large dataset and powerful computational ability. However, most CNN-based methods require training examples with references, whereas it is extremely challenging to simultaneously capture low-light and normal-light images of the same visual scene. To eliminate the reliance on paired training data, several unsupervised deep learning-based methods~\cite{guo2020zero,zhang2020self,jiang2021enlightengan} have been proposed. These algorithms are able to restore images with better illumination and contrast in some cases. However, most unsupervised methods heavily rely on carefully selected multi-exposure training data or unpaired training data, which makes these approaches not generalize well to handle various types of images. Therefore, it is of great interest to seek a novel strategy to deal with different scenarios in the wild.

In this study, we propose an unsupervised low-light image enhancement algorithm based on an effective prior termed histogram equalization prior~(HEP). Our work is motivated by an interesting observation on the pre-trained networks: the feature maps of histogram equalization enhanced image and the ground truth are similar. Intuitively, the feature maps of histogram equalization enhanced images can directly provide abundant texture and luminance information~\cite{geirhos2018imagenet}. We show theoretically and empirically that this generic property of the histogram equalization enhanced image holds for many low-light images; more details are shown in Section~\ref{method}. This inspires us to regularize the feature similarity between the histogram equalization enhanced images and the restored images. 

Following~\cite{Chen2018Retinex}, we split the low-light image enhancement process into two stages: image brightening and image denoising. The first stage decomposes the low-light images into illumination and reflectance maps, and the reflectance maps can be regarded as restored images. We formulate the histogram equalization prior to guiding the training process and add an illumination smoothness loss to suppress the texture and color information in the illumination map. However, according to the derivation based on Retinex theory~\cite{zhang2021beyond}, the reflectance maps are contaminated by noise. To improve the image quality, the second stage works as an enhancer to denoise the reflectance map. In this stage, we propose an unsupervised denoising model based on disentangled representation to remove the noise and generate the final enhanced image. The disentanglement is achieved by splitting the content and noise features in a reflectance map using content encoders and noise encoders. Inspired by~\cite{bao2018towards}, we add a KL divergence loss to regularize the distribution range of extracted noise features to suppress the contained content information. Moreover, we adopt the adversarial loss and the cycle-consistency loss as regularizers to assist the generator networks in yielding more realistic images and preserving the content of the original image. Extensive experiments demonstrate that our method performs favorably against the state-of-the-art unsupervised low-light enhancement algorithms and even matches the state-of-the-art supervised algorithms. Fig.\ref{fig:intro} shows an example of enhancing the low-light image. In comparison to state-of-the-art methods, our method delivers improved image brightness while preserving the details.

In summary, the main contributions of this work are as follows:

1. We propose an effective prior termed histogram equalization prior (HEP) for low-light image decomposition and add an illumination smoothness loss to suppress the texture and color information in the illumination map.

2. We introduce a noise disentanglement module to disentangle the noise and content in the reflectance maps with the reliable aid of unpaired clean images.

3. We build an unsupervised low-light image enhancement framework based on Retinex and disentangled representation, possessing more effective training and faster convergence speed.

4. We demonstrate that the proposed method achieves remarkable performance compared with the state-of-the-art unsupervised algorithms and even matches the state-of-the-art supervised algorithms.

The rest of this paper is organized as follows. Section~\ref{related} provides a brief review of some related works. Section~\ref{method} presents our proposed histogram equalization prior first, then introduces the decomposition network, finally, presents the proposed noise encoder. Section~\ref{experiment} illustrated the experimental results. Section~\ref{ablation} provided the ablation studies on each component. Finally, concluding remarks are provided in Section~\ref{conclusion}.

\section{Related Work}
\label{related}
\textbf{Conventional Methods} The conventional methods for low-light image enhancement can be roughly divided into three aspects: Gamma Correction (GC)~\cite{farid2001blind}, Histogram Equalization (HE)~\cite{pizer1990contrast}, and Retinex~\cite{land1971lightness}. Gamma correction edits the gamma curve of the image to perform nonlinear tone editing to detect the dark part and the light part in the image signal and increase the ratio of the two-part to improve the contrast. However, the global parameters lead to local over/under-exposure, and the value of the global parameter is very complicated to select. Rahman~\etal~\cite{rahman2016adaptive} proposed an adaptive gamma correction method that dynamically determines the intensity conversion function based on the statistical characteristics of the image. 

Histogram Equalization stretches the image's dynamic range by evenly distributing the pixel values to improve the contrast and brightness of the image. However, it applies the adjustment globally, leads to unexpected local overexposure and amplifying the noise. Adaptive Histogram Equalization (AHE)~\cite{pizer1987adaptive} has been proposed to map the histogram of the local region as a simple mathematical distribution. Pizer~\etal~\cite{pizer1990contrast} proposed Contrast Limited Adaptive Histogram Equalization (CLAHE). This method sets a threshold and assumes that if a certain pixel value of the histogram exceeds the threshold, crop this pixel and evenly distribute the part that exceeds the threshold to each pixel. 

Retinex theory is a calculation theory of color constancy. As a model of human visual perception, these methods decompose images into reflectance and illumination maps. MSR~\cite{jobson1997multiscale} obtains enhanced results by fusing different single-scale Retinex outputs. MSRCR~\cite{jobson1997multiscale} improves the color distortion problem of the previous methods. However, the Retinex methods lead to unreal or partially over-enhanced. Inspired by the Retinex theory, NPE~\cite{wang2013naturalness} was proposed for the enhancement of non-uniform illumination images. MF~\cite{fu2016fusion} was proposed to apply multi-layer fusion to image enhancement under different light conditions. LIME~\cite{guo2016lime} evaluate the illumination map of the image and smooth the illumination map for enhancement. SRIE~\cite{fu2016weighted} evaluate the illumination map and the reflectance map simultaneously through a weighted variational model.

\textbf{Deep learning based Methods} Deep learning-based methods have dominated the research of low-light image enhancement. Lore~\etal~\cite{lore2017llnet} proposed the first convolutional neural networks for low-light image enhancement termed LL-Net, perform contrast enhancement and denoising based on deep auto-encoder. Chen~\etal~\cite {Chen2018Retinex} proposed Retinex-Net, which includes a Decom-Net that splits the input images into reflectance and illumination maps, and an Enhance-Net that adjusts the illumination map for low-light enhancement. Zhang~\etal proposed KinD~\cite{zhang2019kindling}, which is similar to Reinex-Net. It presented a new decomposition network, a reflection map enhancement network, and an illumination map enhancement network, which achieved outstanding performance in low-light image enhancement. Zhang~\etal proposed KinD++~\cite{zhang2021beyond}, which improves the KinD method, and achieves state-of-the-art performance. Guo~\etal~\cite{guo2020zero} proposed a zero-shot learning method named Zero-DCE, which is achieved by an intuitive and straightforward nonlinear curve mapping. However, Zero-DCE heavily relies on the usage of multi-exposure training data. Zhang~\etal~\cite{zhang2020self} proposed a self-supervised method that uses the max entropy loss for better image decomposition, but the restored image still suffers from noise contamination.
\begin{figure*}[htbp]
\centering
\includegraphics[width=\textwidth]{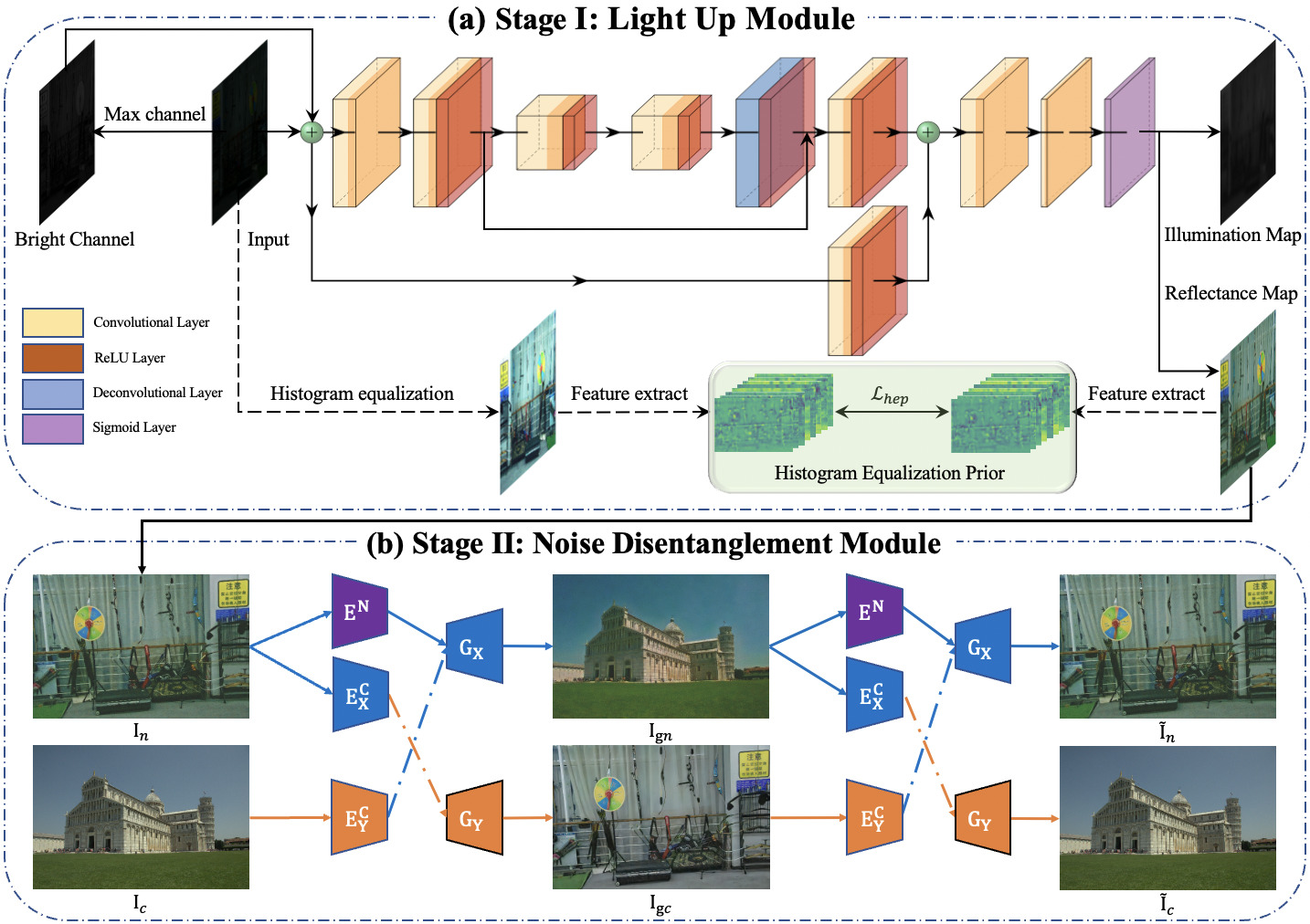}
\caption{Overview of the framework. The proposed method consists of two stages: (a) light-up and (b) noise disentanglement. The light-up module first decomposes the low-light image into an illumination map and reflectance map. Then the noise disentanglement module denoises the reflectance map to yield the final enhanced image. In (a), the bright channel is a 1-channel image, which is obtained by calculating the maximum channel value of the input RGB image. Then the bright channel and the input image are concatenated together to form a 4-channel image as the input of the network. In (b), blue arrows represent the data flow of the noise domain, orange arrows represent the data flow of the clean domain. $E^N$ is the noise encoder for noise images; $E^C_Y$ and $E^C_X$ are the content encoders for noise and clean images; $G_X$ and $G_Y$ are noise image and clean image generators.}
\label{fig arc}
\end{figure*}

\textbf{Image to Image Translation} Generative Adversarial Network~(GAN) is the most influential generative model in computer vision technology. Based on the powerful generative capabilities of GAN, image-to-image translation has become an important way to achieve image enhancement, which is achieved by converting corrupted images to sharp images. Zhu \etal~\cite{zhu2017unpaired} proposed CylceGAN, which showed tremendous capacity in the field of the image domain transfer. Liu \etal~\cite{liu2017unsupervised} proposed UNIT, which learned shared-latent representation for diverse image translation. Lee \etal~\cite{lee2018diverse} proposed DRIT, which separated the latent space to content space and attribute space. The content space is shared, the attribute space is independent. Yuan~\etal~\cite{yuan2018unsupervised} proposed a nested CycleGAN to achieve the unsupervised image super-resolution. Lu~\etal~\cite{lu2019unsupervised} extended DRIT and proposed to decompose the image into the image content domain and the noise domain to achieved unsupervised image deblurring. Based on Lu's work, Du~\etal~\cite{du2020learning} added Background Consistency Module and Semantic Consistency Module to the networks, learning robust representation under dual-domain constraints, such as feature and image domains. Jiang~\etal~\cite{jiang2021enlightengan} proposed a backbone model EnlightenGAN for low-light image enhancement based on adversarial learning. However, EnlightenGAN relies on large number of parameters for good performance.

\section{Methodology}
\label{method}
The main purpose of our method is to recover texture details, reduce noise and color bias, and maintain sharp edges for low-light image enhancement. As shown in Fig.\ref{fig arc}, the proposed method consists of two components: 1) Light Up Module (LUM); 2) Noise Disentanglement Module (LUM). The first stage is improving the brightness of the images, and the second stage is removing the noise of the images.
\begin{figure*}[htbp]
\centering
\includegraphics[width=\textwidth]{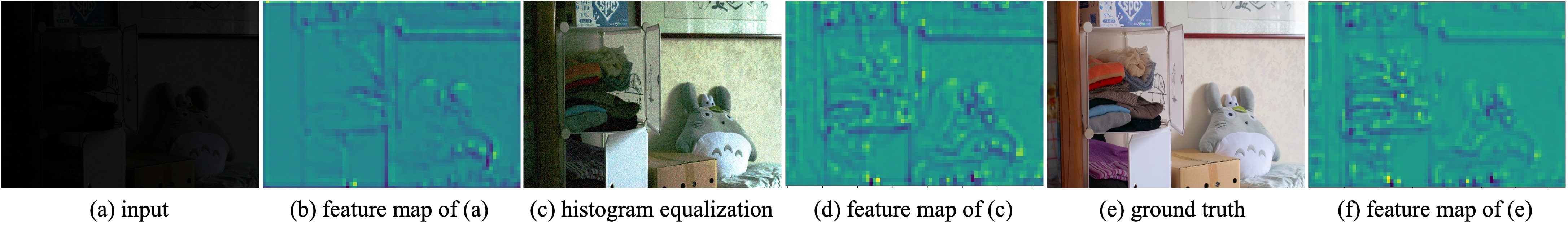}
\caption{Feature maps on the $conv4\_{1}$ layer of VGG-19 networks pre-trained on ImageNet.}
\label{fig prior}
\end{figure*}
\begin{figure}[t]
\centering
 \includegraphics[width=\linewidth]{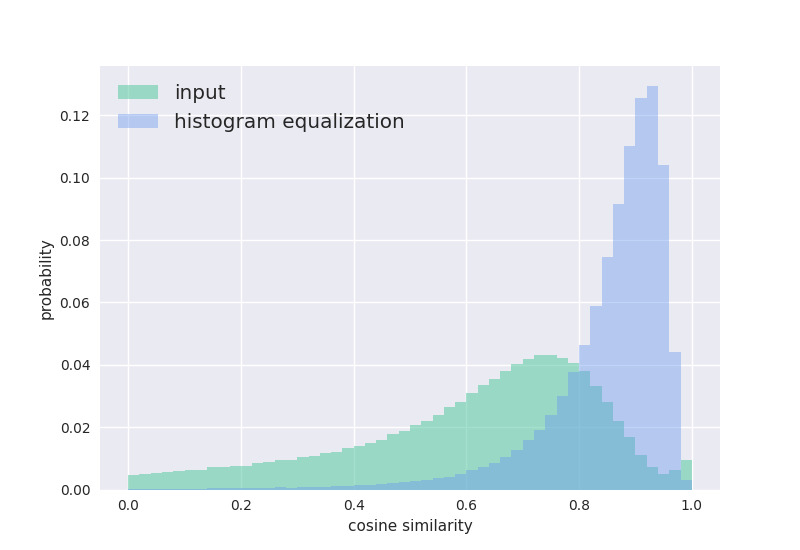}
 \caption{Histogram of the cosine similarities. Green: cosine similarities between the feature maps of the input low-light images and the ground truth. Blue: cosine similarities between the histogram equalization prior and the ground truth.}
 \label{fig cos}
\end{figure}

For low-light image enhancement, unsupervised learning-based methods are complicated to implement. The main reason is that texture and color information in low-light images is difficult to extract without the aid of paired ground truth data or prior information. Therefore, we investigate an effective prior information to guide the training process and maintain the texture and structure. In the following subsections, we first introduce the proposed histogram equalization prior in Section~\ref{hist}. Then, we present the method to decompose the low-light images into reflectance maps and illumination maps in Section~\ref{lightup}. In Section~\ref{noise}, we discuss the approach to disentangle the noise and content in reflectance maps.
\begin{figure}
\centering
 \includegraphics[width=1\linewidth]{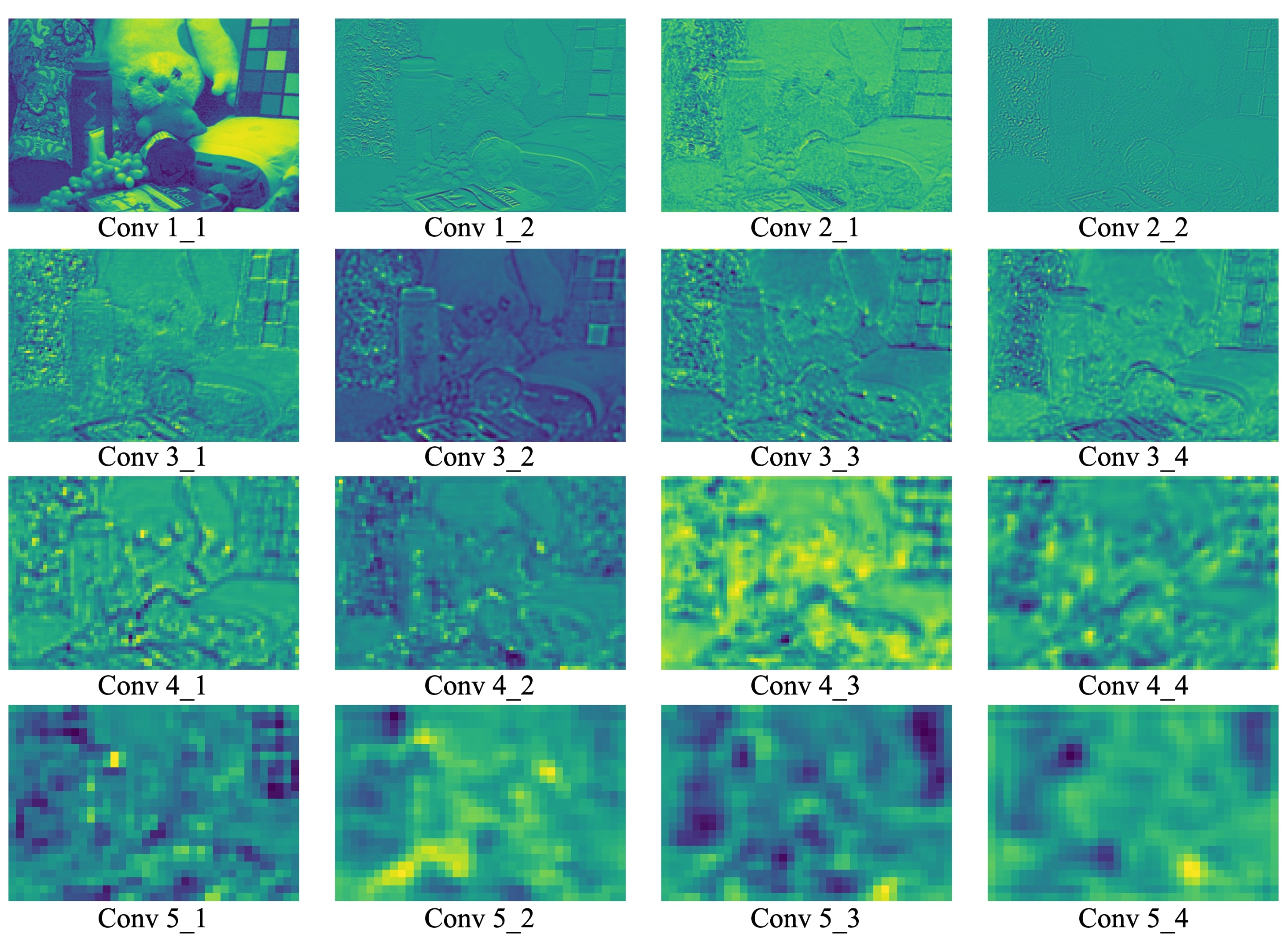}
 \caption{Feature maps on different layers of VGG-19 networks pre-trained on ImageNet with a histogram equalization enhanced image.}
 \label{fig fea}
\end{figure}

\subsection{Histogram Equalization Prior}
\label{hist}
The histogram equalization prior is based on histogram equalization enhanced image. Traditional histogram equalization can make the dark images visible by stretching the dynamic range of dark images via manipulating the corresponding histogram. However, it is not flexible enough for visual property adjustment in local regions and leads to undesirable local appearances, e.g., under/over-exposure and amplified noise. Encouraging the pixels of output images to match the histogram equalization enhanced images will capture unpleasant local impressions contained in the enhanced image. Inspired by \cite{johnson2016perceptual}, we can adopt the VGG feature map to constrain the perceptual similarity between the low-light image and its histogram equalization enhanced version. As shown in Fig.\ref{fig prior}, we can observe that the feature map of the input low-light image has less semantic information~\cite{wang2021rethinking}. In contrast, the feature map of histogram equalization enhanced image has rich semantic information, and it is remarkably similar to the feature map of ground truth.

To further verify the validity of the histogram equalization prior, we have selected 500 paired images from the LOL dataset~\cite{Chen2018Retinex}. We calculate the cosine similarity between the feature maps of the histogram equalization enhanced image and the feature maps of the ground truth. Fig.\ref{fig cos} is the histogram of cosine similarities over all 500 low-light images. We can observe that about $80\%$ of the cosine similarities are concentrated above $0.8$. Compared with the cosine similarities between the feature maps of the input low-light images and the feature maps of the ground truth, the cosine similarities has been substantially improved. This statistic provides a strong support to our histogram equalization prior, and it indicates that we can adopt this prior instead of ground truth to guide the training process.

Fig.\ref{fig fea} shows the different layer of VGG-19~\cite{simonyan2014very} networks pre-trained on ImageNet~\cite{deng2009imagenet} with a histogram equalization enhanced image. Feature maps closer to the input layer pay more attention to the specific details of texture information, and some feature maps can also show the shape of the toy's face. Feature maps farther away from the input layer are more concerned with semantic and abstract information, such as the toy's eye and nose characteristics. The feature maps of the deepest layers become more obscure and can no longer provide adequate information, while the features are similar between each group of feature maps. Based on this information, we select the feature map of $conv4\_{1}$ layer as the feature similarity.

\subsection{Light Up}
\label{lightup}
The first stage improves the brightness of the images based on the Retinex theory. According to the Retinex theory, the images can be decomposed into reflectance maps and illumination maps. Mathematically, a degraded low-light image can be naturally modeled as follows:
\begin{equation}
    I=R \circ L + N
\end{equation}
\noindent where I stands for input image, R stands for reflectance map, L is the illumination map, N represents the noise component, $\circ$ represents element-wise multiplication. 

As the illumination map determines the dynamic range of images, it cannot be affected by noise. In contrast, the reflectance map represents the intrinsic properties of the images, which are often affected by noise during the imaging process. Hence, by taking simple algebra steps~\cite{zhang2021beyond}, we can have the following formula:
\begin{equation}
I=R \circ L + N=R \circ L + \tilde{N} \circ L =(R + \tilde{N}) \circ L=\tilde{R} \circ L
\end{equation}
\noindent where $\tilde{N}$ stands for the degradation having the illumination decoupled, $\tilde{R}$ represents for the polluted reflectance map. 

According to the above theory, the reflectance map can be regarded as a restored image with noise. Therefore, we design a neural network to decompose the low-light images into reflectance and illumination maps, and then send the reflectance maps to the NDM for further denoising. We follow the similar network architecture as the one used in~\cite{zhang2020self}, the module framework is shown in Fig.\ref{fig arc}(a). It first uses a 9×9 convolutional layer to extract features from the input image. Secondly, three 3×3 convolutional+ReLU layers and one 3×3 deconvolutional+ReLU layer are followed. A residual feature from the conv2 layer concatenates with the feature from the deconv layer and feeds to a 3×3 convolutional+ReLU layer. The feature from this layer concatenates with the feature from a 3×3 convolutional+ReLU layer, which extracts features from the input image. Finally, two 3×3 convolutional layers project reflectance map and illumination map from feature space. The sigmoid function constrains both reflectance map and illumination map in the range of [0,1].

Due to the lack of ground-truth data to guide the training process, it is tough to recover these two components from low-light images. We adopt the histogram equalization prior to constrain the reflectance map. We define an MSE loss between the feature map of the output reflectance map and the feature map of the input image, which we call the histogram equalization prior loss. The loss function can be formulated as follows:
\begin{equation}
\mathcal{L}_{hep} = \parallel F(\tilde{R}) - F(I)\parallel_2^2
\end{equation}
\noindent where $F(\cdot)$ denotes the feature map extracted from a VGG-19 model pre-trained on ImageNet.

Since the network decomposes the image into an illumination map and a reflectance map, the decomposed two maps should reproduce the input image. We introduce reconstruction loss to ensure the quality of the generated image. The formula is as follows:
\begin{equation}
\mathcal{L}_{recon} = \parallel \tilde{R} \circ L - I\parallel_1
\end{equation}

As the reflectance map should preserve more texture and color details. In other words, the illumination map should be smooth in textural information while still preserving the structural boundaries. To make the illumination map aware of the image structure boundary, we modify the illumination smoothness loss proposed in \cite{zhang2021beyond}. Different from the previous loss, our illumination smoothness loss only takes the low-light input image as the reference. This term constrains the relative structure of the illumination map to be consistent with the input image, which can reduce the risk of over-smoothing on the structure boundary. The illumination smoothness loss is formulated as:
\begin{equation}
\mathcal{L}_{is} = \parallel \frac{\nabla L}{max(\mid \nabla I \mid, \epsilon)}\parallel_1
\end{equation}
\noindent where $\mid\!\cdot\!\mid$ means the absolute value operator, $\epsilon$ is a small positive constant for avoiding zero denominators, $\nabla$ denotes the gradient including $\nabla h$ (horizontal) and $\nabla v$ (vertical).

As a result, the loss function of the LUM is as follows:
\begin{equation}
\mathcal{L} = \mathcal{L}_{recon} + \lambda_{rs}\mathcal{L}_{hep} + \lambda_{is}\mathcal{L}_{is}
\end{equation}


In our experiment, these parameters are set to $\lambda_{hep}=\lambda_{is}= 0.1$, $\epsilon=0.01$. Due to the careful settings of these loss terms, the light-up module can perform sufficiently well. Still, the light-up image is constrained by histogram equalization, as the method often causes noise and blur. Although the images generated by the network have been enhanced, however, compared with the normal-light images, the noise level cannot meet the visual quality. Therefore, they need to be further denoised.

\subsection{Noise Disentanglement}
\label{noise}
Although the content information of the low-light image appears after being decomposed into a reflectance map, the noise contained in it seriously interferes with the clarity of the image. Therefore, we adopt the domain transfer method to eliminate the noise and retain the content information. As shown in Fig.\ref{fig arc}(b). The noise disentanglement module consists of six parts: 1) content encoder $E_X^C$ and $E_Y^C$(due to the shared parameter, we regard the content encoder of two domains are the same); 2) noise encoder $E^N$; 3) noise domain image generator $G_X$; 4) clean domain image generator $G_Y$; 5) noise domain image discriminator $D_X$; 6) clean domain image discriminator $D_Y$. Given a training image sample $I_n$ from the noise domain and a training image sample $I_c$ from the clean domain, the content encoder $E_X^C$ and $E_Y^C$ extract the content feature from corresponding image samples, the noise encoder $E^N$ extract the noise feature from noise image samples. $G_X$ then takes the content feature of the clean domain and noise feature of the noise domain to generate a noise image $I_{gn}$ while $G_Y$ then takes the content feature of the noise domain to generate a clean image $I_{gc}$. The discriminators $D_X$ and $D_Y$ distinguish between the real and generated examples.

Due to the unpaired setting, it is not trivial to disentangle the content information from a noise image. To restrain the noise encoder to only encode the noise information, we add the KL distance to constrain the distribution of noise feature extracted by the noise encoder, forcing the distribution of the noise feature to be closer to the standard normal distribution. The KL distance formula is as follows:
\begin{equation}
KL(q(z_n)\parallel p(z))=\int q(z_n)\log \frac{p(z)}{q(z_n)}dz
\end{equation}
\noindent where $q(z_n)$ stands for the distribution of the noise features $z_n$, $p(z)$ stands for the distribution of the standard normal distribution $N(0,1)$. 

As proved in \cite{kingma2013auto}, the KL divergence loss will suppress the content information contained in the noise feature $z_n$, and minimizing the KL divergence is equivalent to minimizing the following loss function, which has been proved in \cite{bao2018towards}.
\begin{equation}
\mathcal{L}_{KL}=\frac{1}{2}\sum_{i=1}^d(-\log(\sigma_i^2)+\mu_i^2+\sigma_u^2-1)
\end{equation}
\noindent where $d$ is the dimension of noise feature, $\mu$ and $\sigma$ are the mean and standard deviation of noise feature. 

In order to make the enhanced images look like realistic normal-light images, we adopt the adversarial loss to minimize the distance between the real image and output distributions. We modified the discriminator slightly to replace the loss function with the least-square GAN (LSGAN) loss. The adversarial loss function is as follows:
\vspace{-1mm}
\begin{equation}
\label{adv_loss_x}
\mathcal{L}_{adv}=\frac{1}{2}\mathbb{E}_{x\sim p_r}[(D(x)-b)^2]+\frac{1}{2}\mathbb{E}_{z\sim p_z}[(D(G(z))-a)^2]
\end{equation}
\noindent where $a$ is the label for the generated samples, $b$ is the label for the real samples, and $z$ is the latent vector.

Without pairwise supervision, the denoised image may lose some content information. Similar to~\cite{zhu2017unpaired}, we introduce the cycle-consistency loss to guarantee that the generated corrupted image $I_{gn}$ translates back to the original clean image domain, and the denoised image $I_{gc}$ reconstructed to the original corrupted sample. We define the cycle-consistency loss on both domains as:
\begin{equation}
\mathcal{L}_{cc}=\parallel I - \tilde{I}\parallel_1
\end{equation}
\noindent where $I$ is the input samples, $\tilde{I}$ is the the backward translation of the input samples.

In addition to the cycle-consistency loss, we introduce self-reconstruction loss to facilitate the better-perceived quality of the generated image. The formula of the loss function is as follows:
\begin{equation}
\mathcal{L}_{rec} = \parallel I_{rec} - I\parallel_1
\end{equation}

Following the observations from \cite{taigman2016unsupervised} that features extracted from the deep layer of pre-trained model contain rich semantic information, we add perceptual loss between the denoised images and the original corrupted images to recover finer image texture details. It could be formulated as:
\begin{equation}
\mathcal{L}_{per}=\parallel \phi_l(I_g) - \phi_l(I)\parallel_2^2
\end{equation}
\noindent where $\phi_l(\cdot)$ represents the feature extracted from $l$-th layer of the pre-trained VGG network, $I_g$ is the generated samples. In our experiments, we use the $conv3\_{2}$ layer of the VGG-19 pre-trained network on ImageNet.

To eliminate the potential color deviations in the denoised image, we adopt the color constancy loss proposed in \cite{guo2020zero}, it follows the Gray-World color constancy hypothesis that color in each sensor channel averages to gray over the entire image. The loss function can be expressed as:
\begin{equation}
    \mathcal{L}_{col}\!=\!\sum\nolimits_{\forall(p,q)\in\epsilon}\! (\mathcal{J}^{p}-\mathcal{J}^{q})^2,\!\epsilon=\{(R,G),(R,B),(G,B)\}
\end{equation}
\noindent where $\mathcal{J}^{p}$ represents the the average intensity value of $p$ channel in the denoised image, $(p,q)$ represents a pair of channels.

From our preliminary experiments, we find that the generated denoised samples often over-smooth in the background, then we adopt background consistency loss proposed by \cite{du2020learning}, which uses a multi-scale Gaussian-Blur operator to obtain multi-scale features respectively. The loss function is formulated as:
\begin{equation}
\mathcal{L}_{bc}=\sum_{\sigma=i,j,k} \lambda_\sigma \parallel B_\sigma(I) - B_\sigma(I_g)\parallel_1
\end{equation}
\noindent where $\lambda_\sigma$ is the hyper-parameter to balance the errors at different Gaussian-Blur levels, $B_\sigma(\cdot)$ represents the Gaussian-Blur operator with blur kernel $\sigma$. In our experiments, we set $\lambda_\sigma={0.25, 0.5, 1.0}$ for $\sigma={5,9,15}$ respectively.

The entire loss function for the NDM is summarized as follows:
\begin{equation}
\begin{split}
\mathcal{L}=& \quad\mathcal{L}_{adv}+\lambda_{KL}\mathcal{L}_{KL}+\lambda_{cc}\mathcal{L}_{cc}+\lambda_{col}\mathcal{L}_{col}\\
& \quad+\lambda_{per}\mathcal{L}_{per}+\lambda_{bc}\mathcal{L}_{bc}+\lambda_{rec}\mathcal{L}_{rec}
\end{split}
\end{equation}

We empirically set these parameters to $\lambda_{KL}=0.01$, $\lambda_{per}=0.1$, $\lambda_{col}=0.5$, $\lambda_{bc}=5$, $\lambda_{cc}=\lambda_{rec}=10$. At test time, given a test corrupted sample, $E^N$ and $E_X^C$ extract the noise and content features map respectively. Then $G_Y$ takes the latent vector and generates the denoised image as the outputs.
\begin{figure*}
    \begin{center}
    \hspace*{-4mm}
    \subfigure[Input]{
        \begin{tabular}[]{c}
        \includegraphics[width=0.19\linewidth]{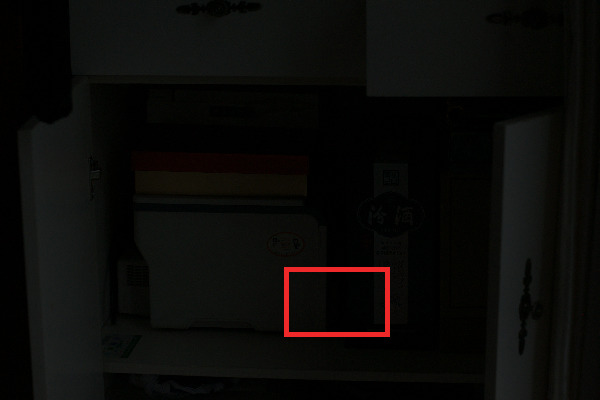}\\
        \includegraphics[width=0.19\linewidth]{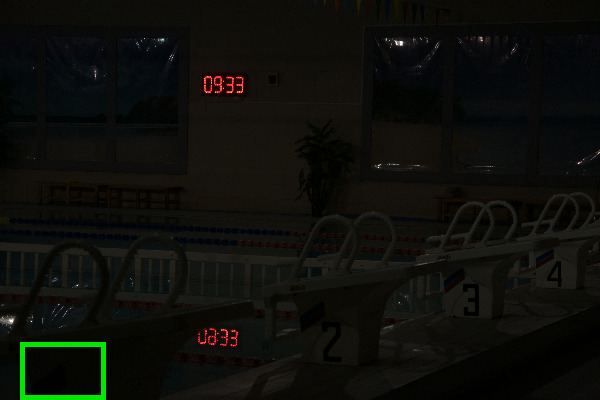}\\
        \includegraphics[width=0.092\linewidth]{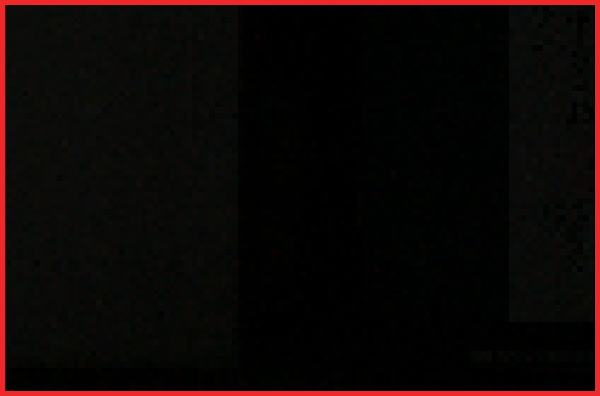}
        \includegraphics[width=0.09\linewidth]{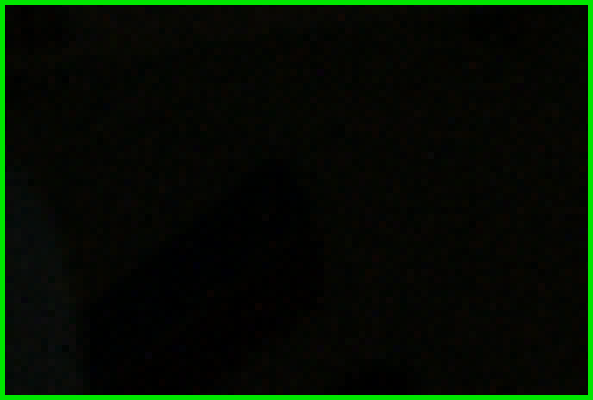}
        \end{tabular}
    }\hspace*{-5mm}
    \subfigure[HE~\cite{pizer1990contrast}]{
        \begin{tabular}[]{c}
        \includegraphics[width=0.19\linewidth]{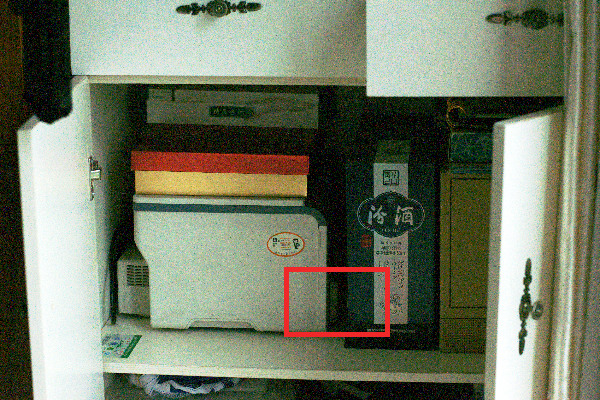}\\
        \includegraphics[width=0.19\linewidth]{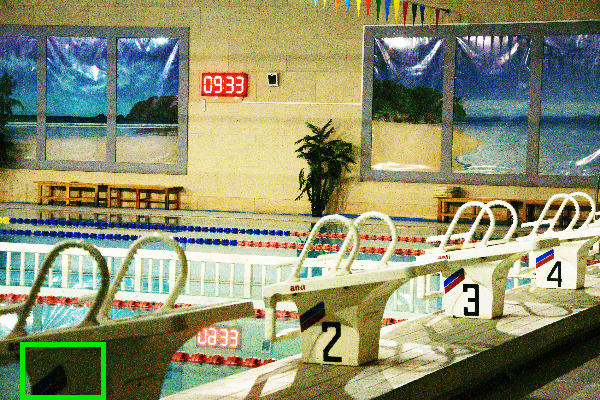}\\
        \includegraphics[width=0.092\linewidth]{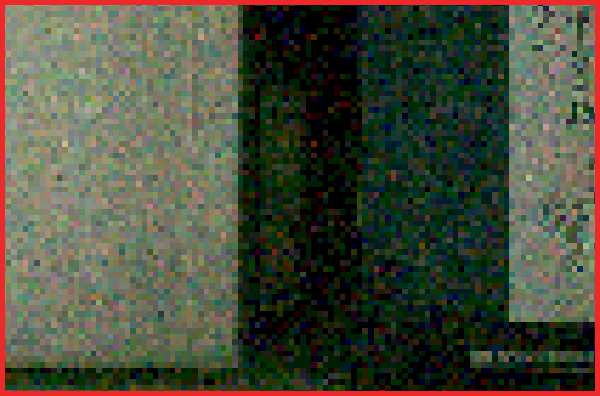}
        \includegraphics[width=0.09\linewidth]{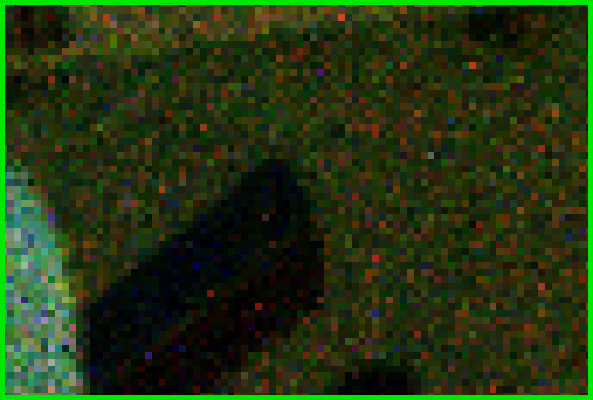}
        \end{tabular}
    }\hspace*{-5mm}
    \subfigure[LIME~\cite{guo2016lime}]{
        \begin{tabular}[]{c}
        \includegraphics[width=0.19\linewidth]{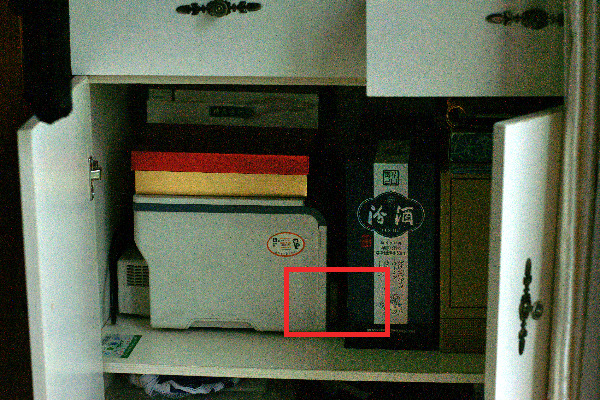}\\
        \includegraphics[width=0.19\linewidth]{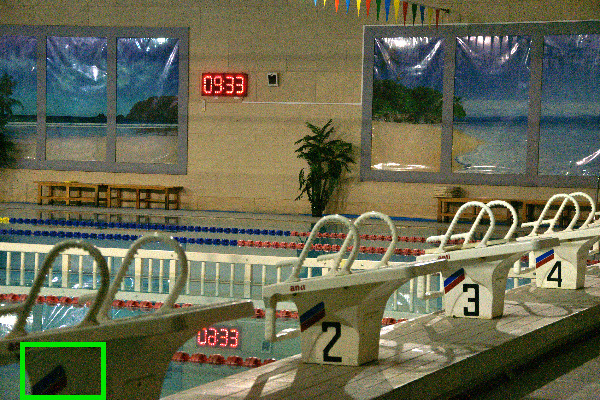}\\
        \includegraphics[width=0.092\linewidth]{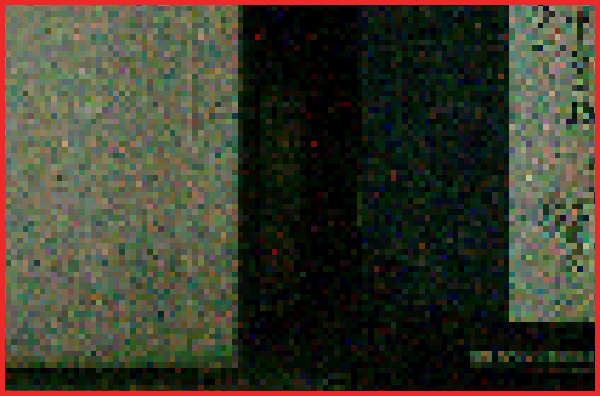}
        \includegraphics[width=0.09\linewidth]{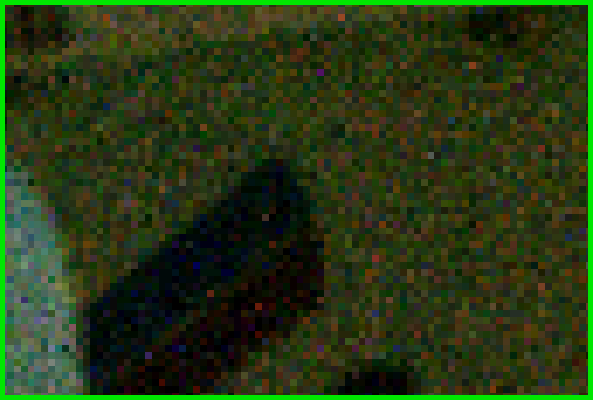}
        \end{tabular}
    }\hspace*{-5mm}
    \subfigure[Retinex-Net~\cite{Chen2018Retinex}]{
        \begin{tabular}[]{c}
        \includegraphics[width=0.19\linewidth]{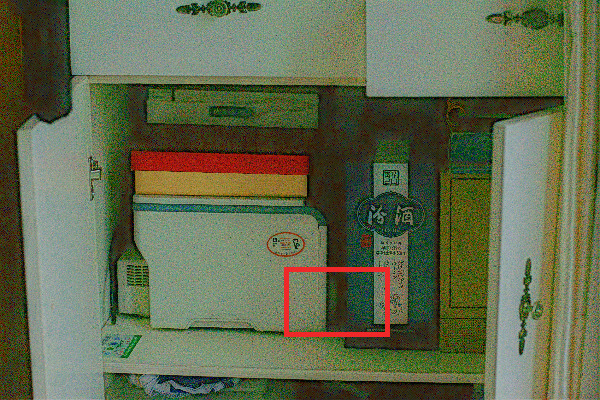}\\
        \includegraphics[width=0.19\linewidth]{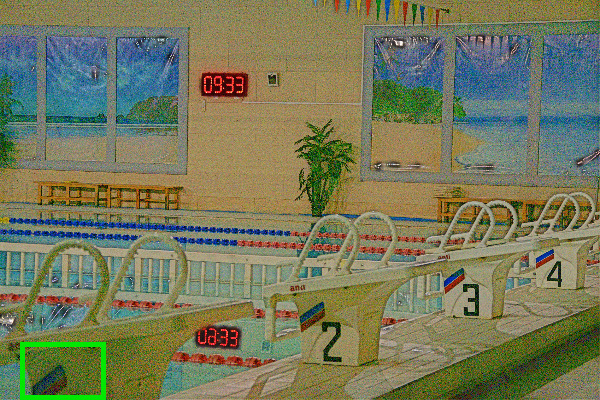}\\
        \includegraphics[width=0.092\linewidth]{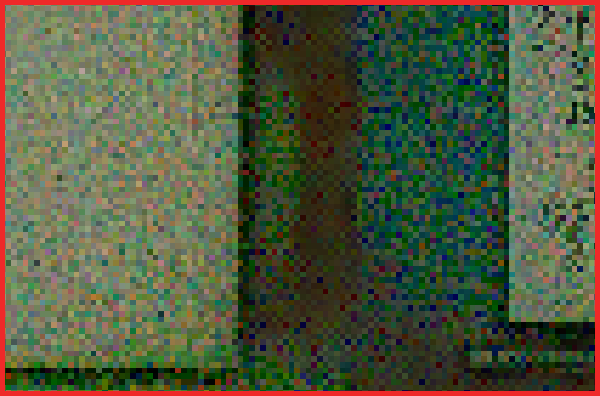}
        \includegraphics[width=0.09\linewidth]{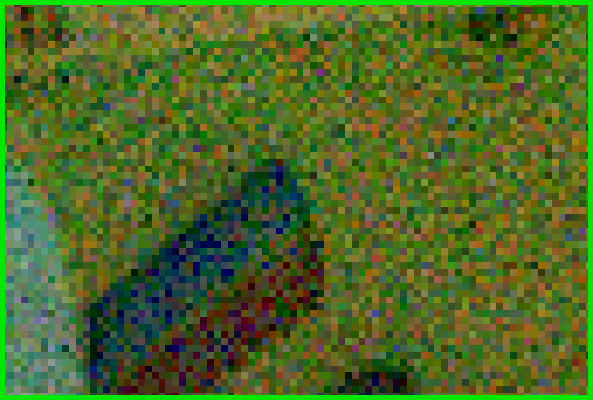}
        \end{tabular}
    }\hspace*{-5mm}
    \subfigure[KinD++~\cite{zhang2021beyond}]{
        \begin{tabular}[]{c}
        \includegraphics[width=0.19\linewidth]{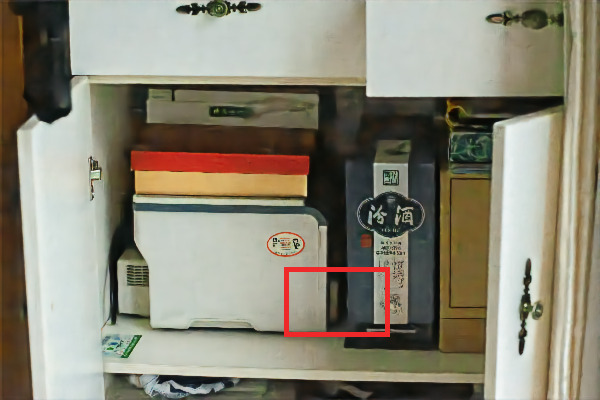}\\
        \includegraphics[width=0.19\linewidth]{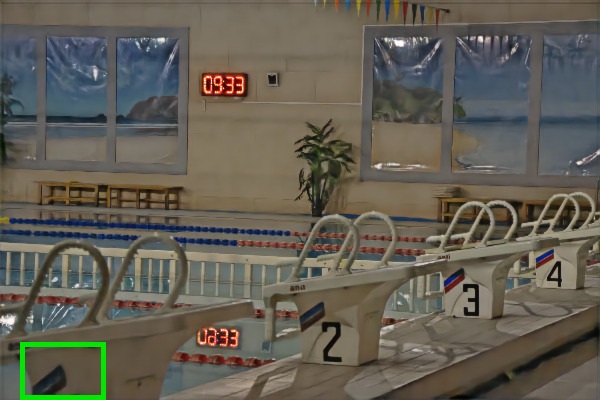}\\
        \includegraphics[width=0.092\linewidth]{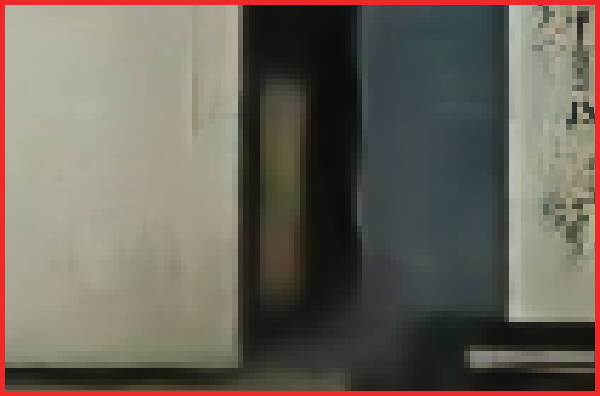}
        \includegraphics[width=0.09\linewidth]{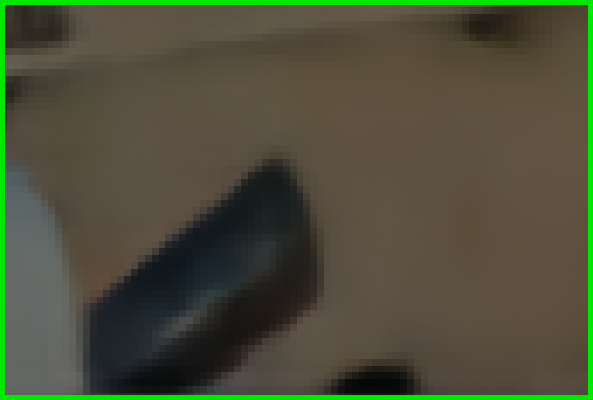}
        \end{tabular}
    }
    \hspace*{-4mm}
    \subfigure[Zero-DCE~\cite{guo2020zero}]{
        \begin{tabular}[]{c}
        \includegraphics[width=0.19\linewidth]{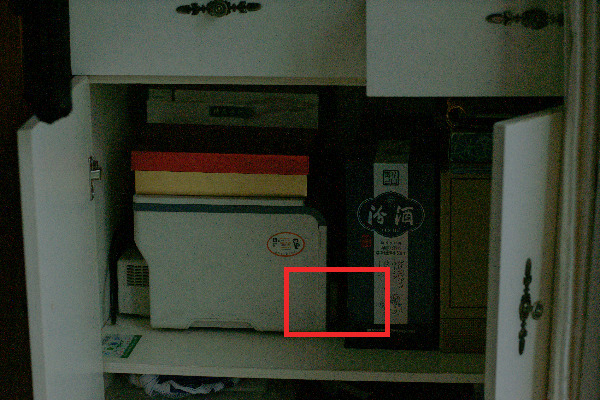}\\
        \includegraphics[width=0.19\linewidth]{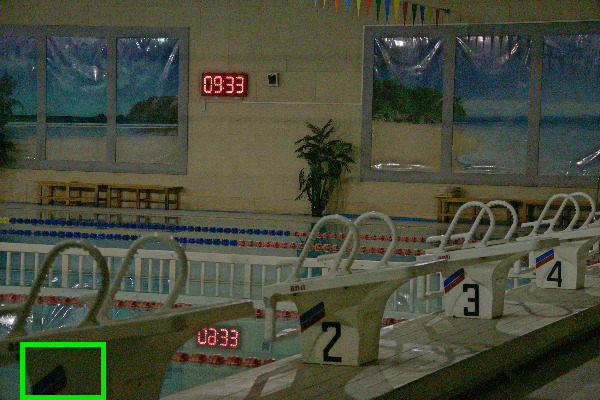}\\
        \includegraphics[width=0.092\linewidth]{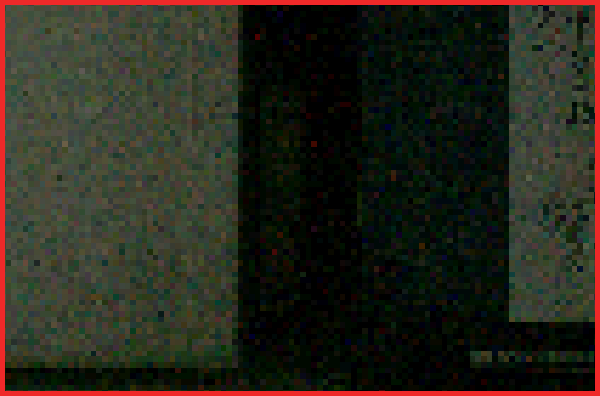}
        \includegraphics[width=0.09\linewidth]{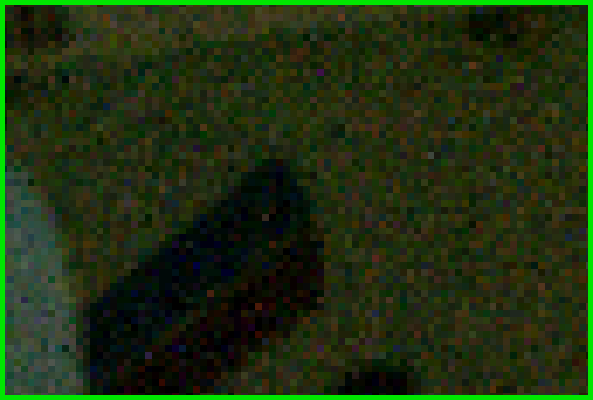}
        \end{tabular}
    }\hspace*{-5mm}
    \subfigure[EnlightenGAN~\cite{jiang2021enlightengan}]{
        \begin{tabular}[]{c}
        \includegraphics[width=0.19\linewidth]{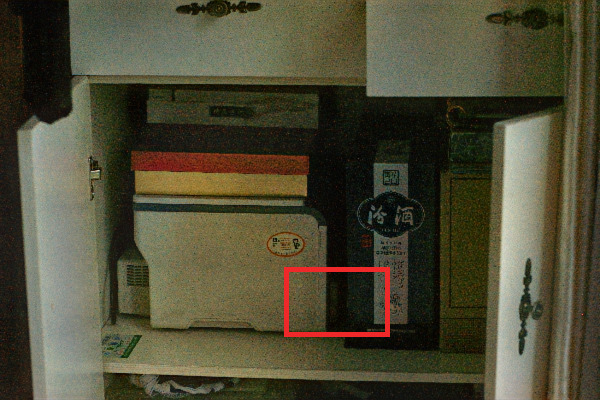}\\
        \includegraphics[width=0.19\linewidth]{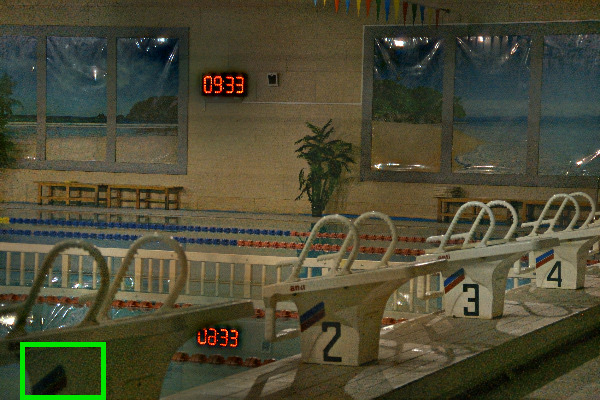}\\
        \includegraphics[width=0.092\linewidth]{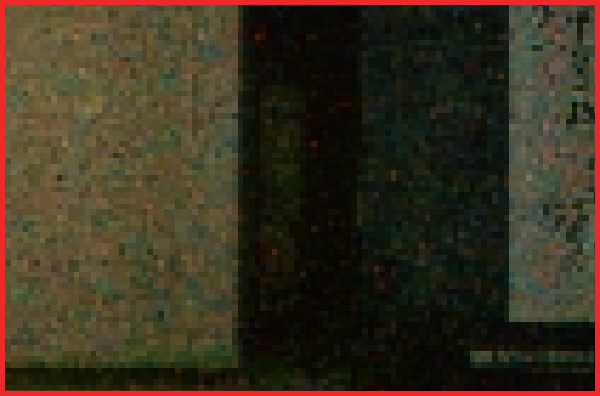}
        \includegraphics[width=0.09\linewidth]{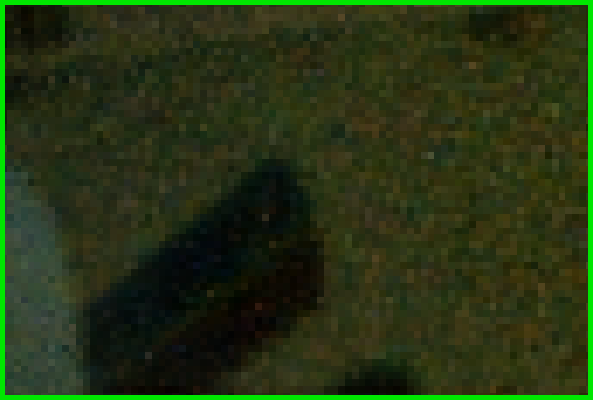}
         \end{tabular}
    }\hspace*{-5mm}
    \subfigure[Self-Supervised~\cite{zhang2020self}]{
        \begin{tabular}[]{c}
        \includegraphics[width=0.19\linewidth]{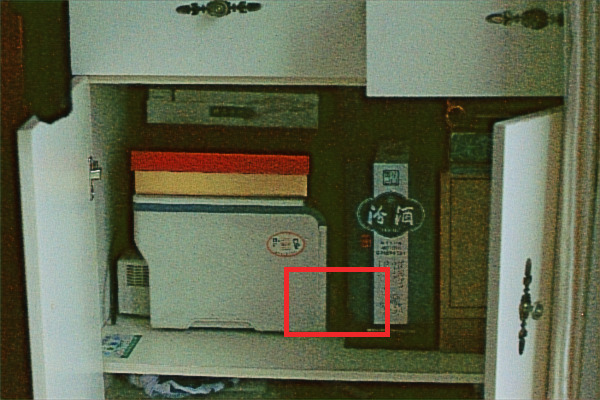}\\
        \includegraphics[width=0.19\linewidth]{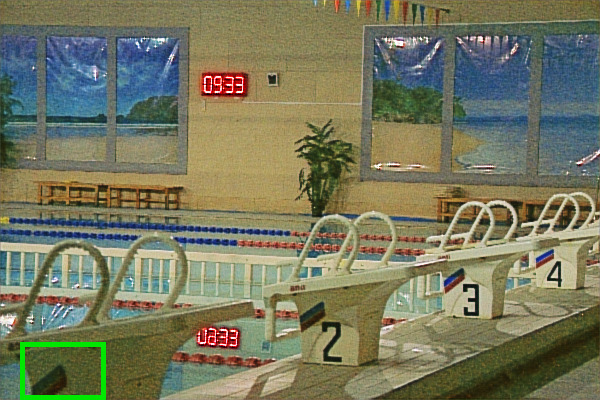}\\
        \includegraphics[width=0.092\linewidth]{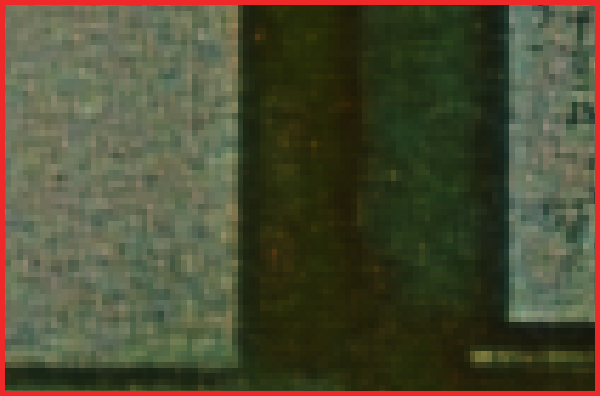}
        \includegraphics[width=0.09\linewidth]{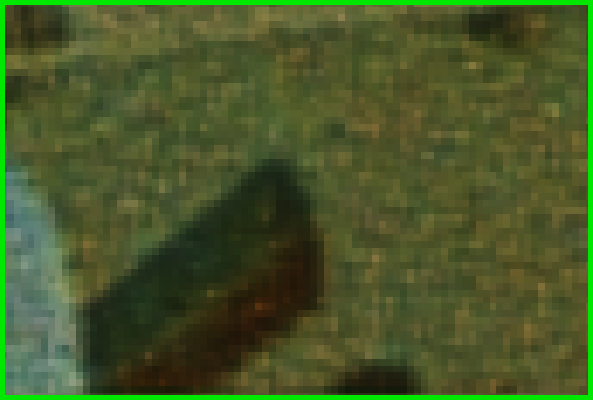}
        \end{tabular}
    }\hspace*{-5mm}
    \subfigure[Ours]{
        \begin{tabular}[]{c}
        \includegraphics[width=0.19\linewidth]{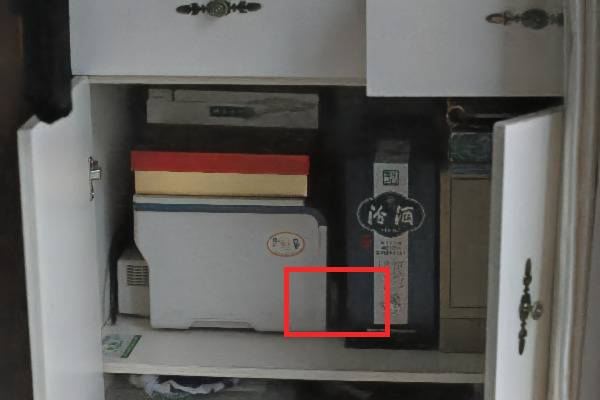}\\
        \includegraphics[width=0.19\linewidth]{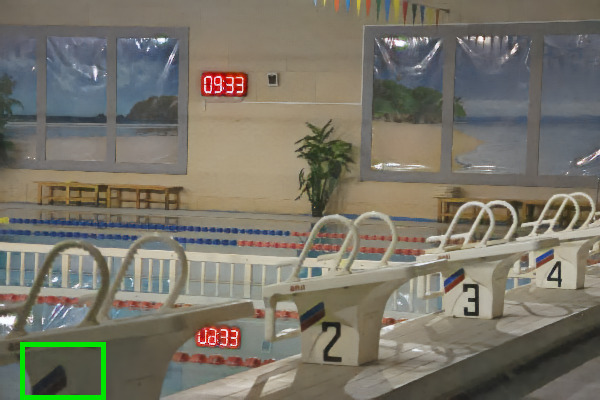}\\
        \includegraphics[width=0.092\linewidth]{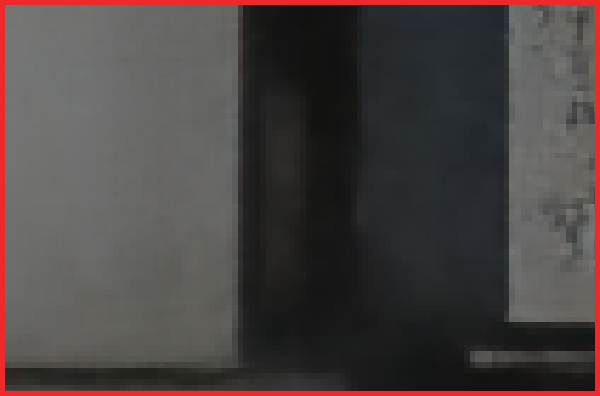}
        \includegraphics[width=0.09\linewidth]{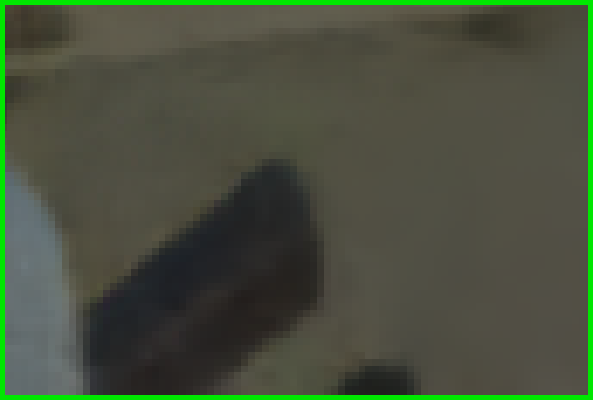}
        \end{tabular}
    }\hspace*{-5mm}
    \subfigure[Ground-Truth]{
        \begin{tabular}[]{c}
        \includegraphics[width=0.19\linewidth]{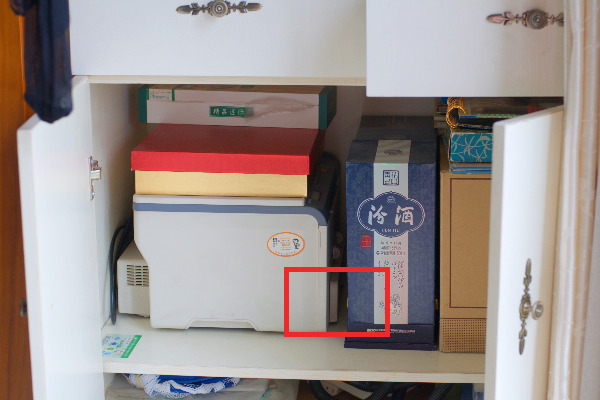}\\
        \includegraphics[width=0.19\linewidth]{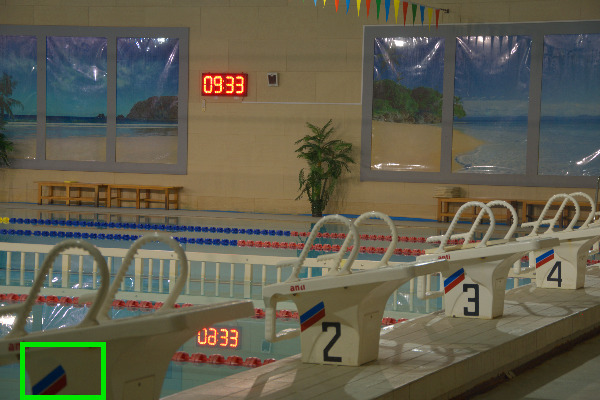}\\
        \includegraphics[width=0.092\linewidth]{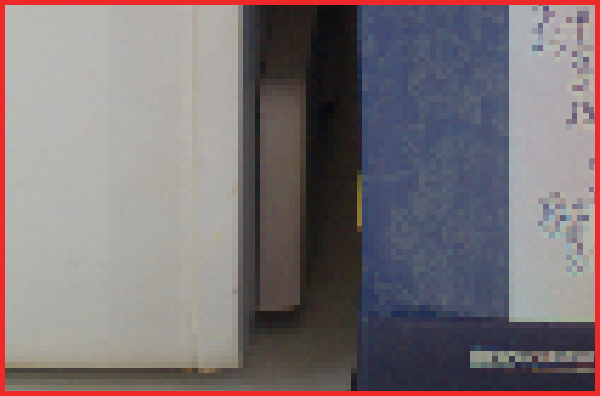}
        \includegraphics[width=0.09\linewidth]{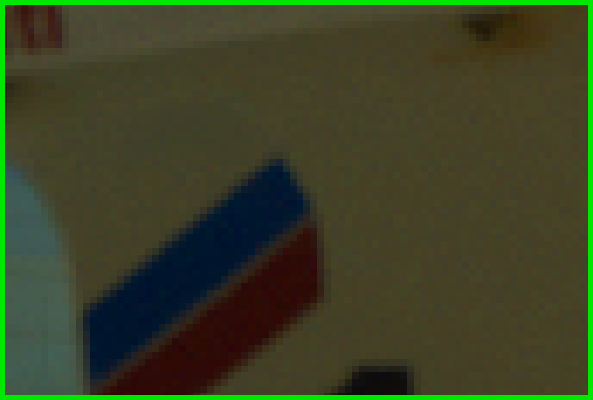}
        \end{tabular}
    }
    \caption{Visual comparison with other state-of-the-art methods on LOL dataset~\cite{Chen2018Retinex}. Best viewed in color and by zooming in.}
    \label{fig:LOL}
    \end{center}
\end{figure*}

\section{Experimental Validation}
\label{experiment}
In this section, we first introduce the implementation details of the proposed method for low-light image enhancement. Then we qualitatively and quantitatively compare the proposed method with the state-of-the-art methods (include supervised and unsupervised methods), we use traditional metrics to evaluate, such as Peak-Signal-Noise-Rate (PSNR), Structural Similarity (SSIM)~\cite{wang2004image}, and Natural Image Quality Evaluator (NIQE)~\cite{mittal2012making}. Furthermore, we test the proposed method on some real-world datasets while comparing them with the state-of-the-art methods in terms of visual performance and NIQE metrics. Finally, we conduct ablation studies to demonstrate the effectiveness of each component or loss in the proposed method.

\subsection{Implementation Details}
Since the proposed method is a two-stage model, we need to train the model separately. In the first stage, our training dataset is selected from the low-light part of the LOL dataset~\cite {Chen2018Retinex}, which includes 500 low/normal-light image pairs. During the training, we use Adam~\cite{kingma2014adam} optimizer to perform optimization with the weight decay equal to 0.0001. The initial learning rate is set to $10^{-4}$, which decreases to $10^{-5}$ after 20 epochs and then to $10^{-6}$ after 40 epochs. The batch size is set to 16 and the patch size to 48×48. In the second stage, we assemble a mixture of 481 low-light images from the LOL dataset and 481 normal-light images from the EnlightenGAN dataset~\cite{jiang2021enlightengan}. The Adam method is adopted to optimize the parameters with the momentum equal to 0.9 and the weight decay equal to 0.0001. The learning rate is initially set to $10^{-4}$ and exponential decay over the 10K iterators. The batch size is set to 16 and the patch size to 64×64. All experiments are conducted using PyTorch~\cite{paszke2017automatic} framework on an Nvidia 2080Ti GTX GPU.
\begin{figure*}
    \begin{center}
    \subfigure[Input]{ 
        \includegraphics[width=0.32\linewidth]{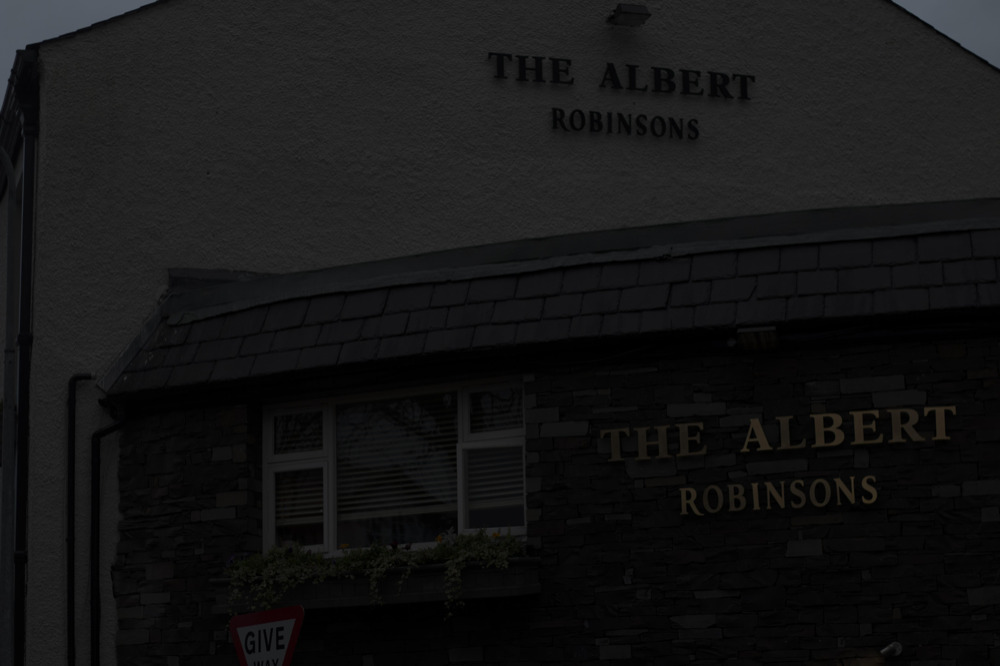}
    }\hspace*{-2mm}
    \subfigure[HE~\cite{pizer1990contrast}]{
        \includegraphics[width=0.32\linewidth]{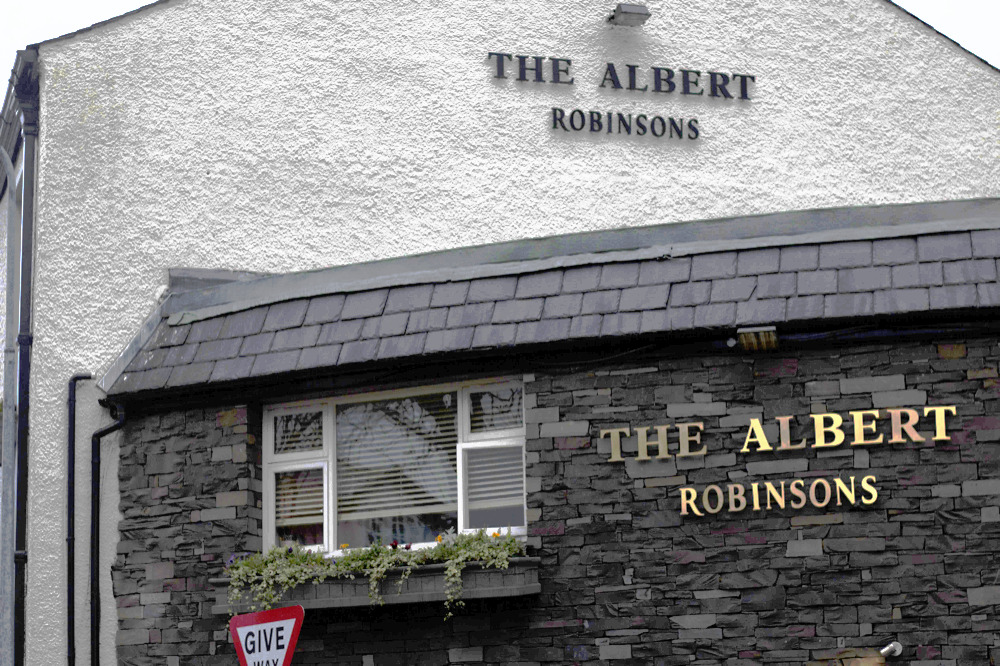}
    }\hspace*{-2mm}
    \subfigure[LIME~\cite{guo2016lime}]{
        \includegraphics[width=0.32\linewidth]{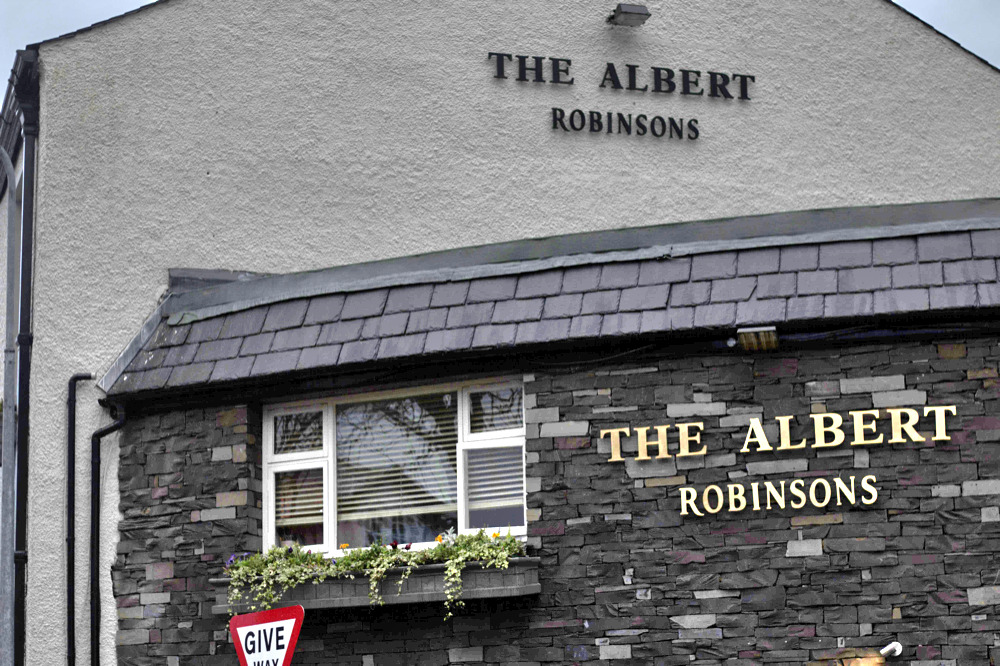}
    }
    \subfigure[Retinex-Net~\cite{Chen2018Retinex}]{
        \includegraphics[width=0.32\linewidth]{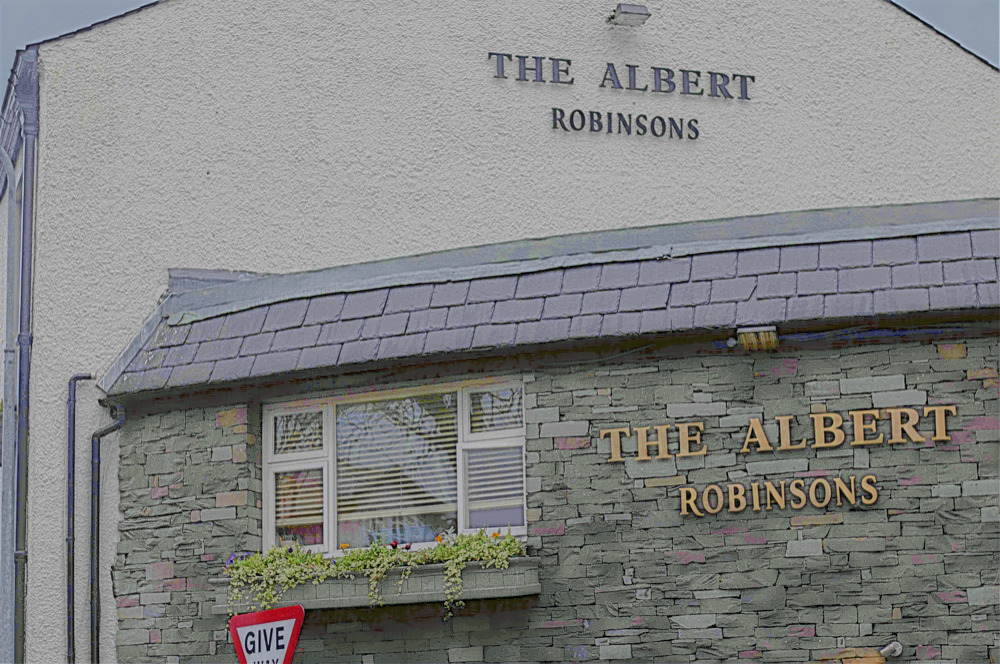}
    }\hspace*{-2mm}
    \subfigure[KinD++~\cite{zhang2021beyond}]{
        \includegraphics[width=0.32\linewidth]{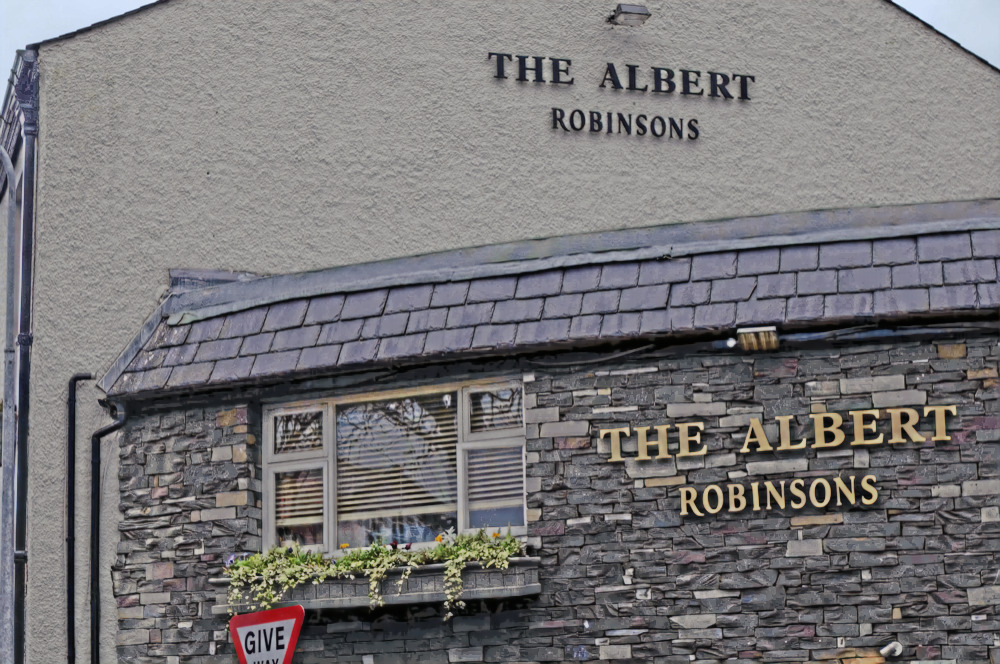}
    }\hspace*{-2mm}
    \subfigure[Zero-DCE~\cite{guo2020zero}]{
        \includegraphics[width=0.32\linewidth]{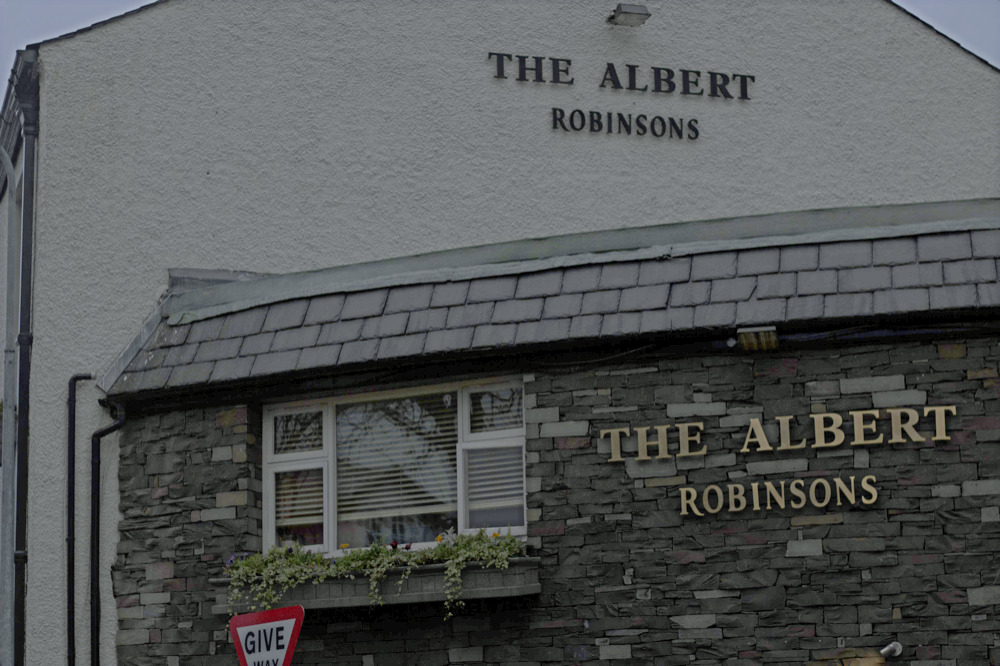}
    }
    \subfigure[EnlightenGAN~\cite{jiang2021enlightengan}]{
        \includegraphics[width=0.32\linewidth]{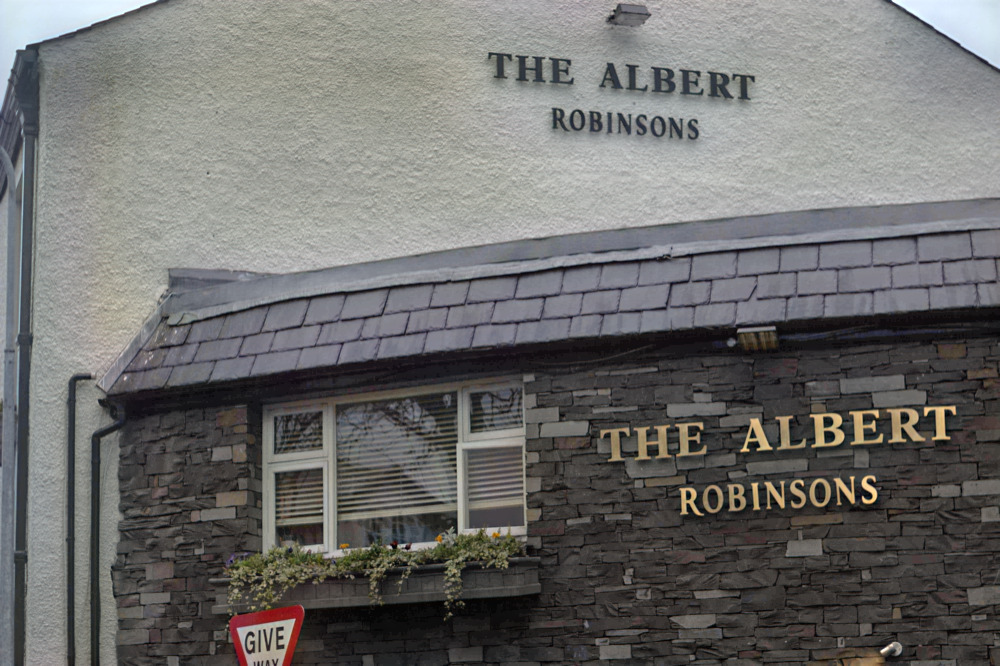}
    }\hspace*{-2mm}
    \subfigure[Self-Supervised~\cite{zhang2020self}]{
        \includegraphics[width=0.32\linewidth]{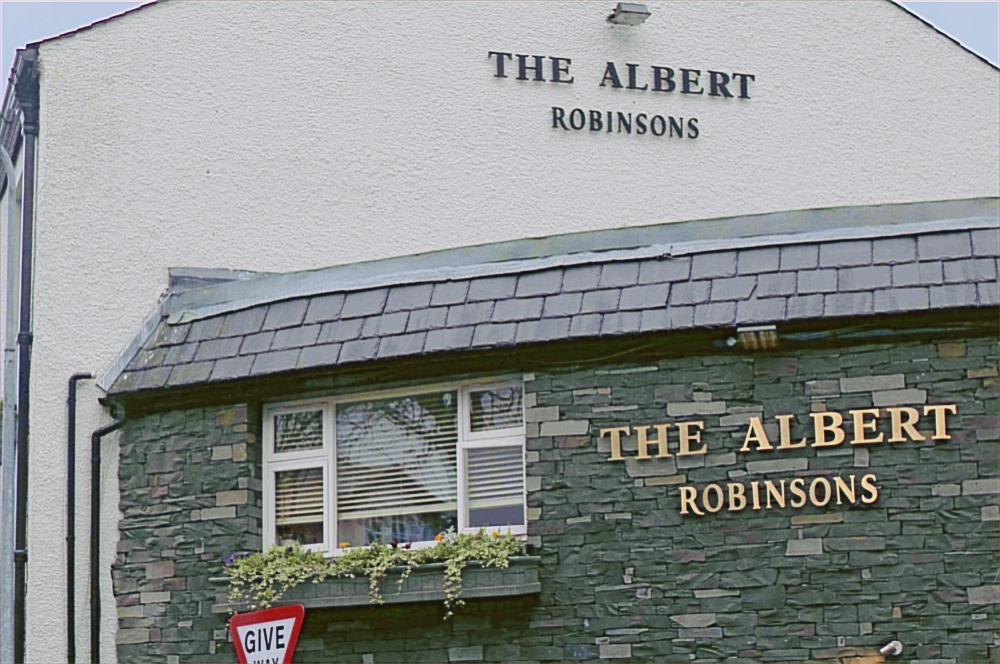}
    }\hspace*{-2mm}
    \subfigure[Ours]{
        \includegraphics[width=0.32\linewidth]{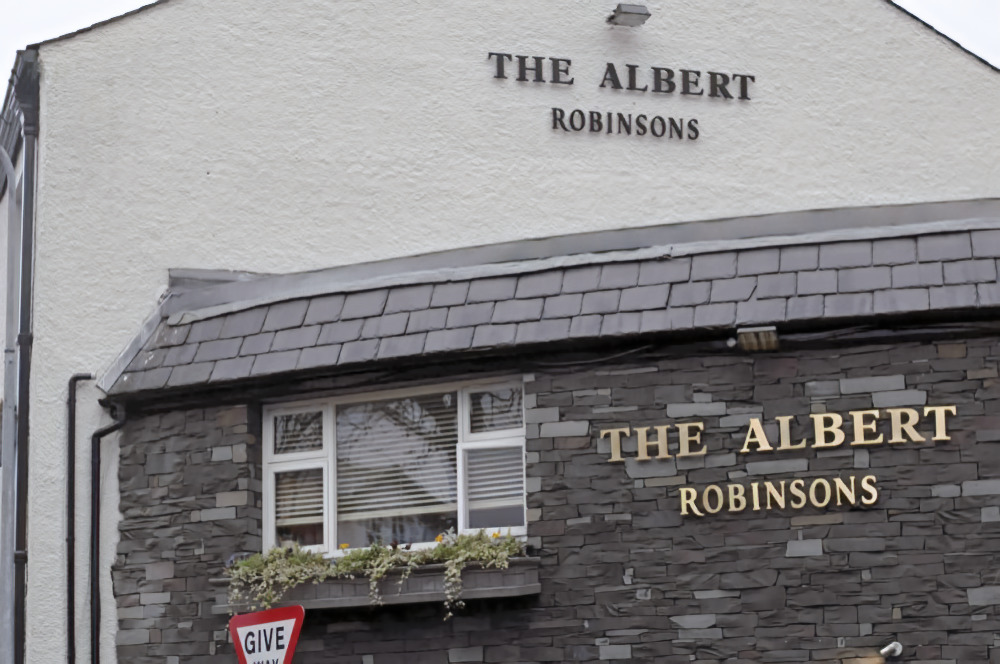}
    }
    \caption{Visual comparison with state-of-the-art methods on the SCIE dataset~\cite{cai2018learning}. Best viewed in color and by zooming in.}
    \label{fig:real}
    \end{center}
\end{figure*}

\subsection{Qualitative Evaluation}
We first visually evaluate our proposed networks on the classical low-light image datasets: LOL datasets, and compare it with other state-of-the-art approaches with available codes, including HE~\cite{pizer1990contrast}, LIME~\cite{guo2016lime}, Retinex-Net~\cite {Chen2018Retinex}, KinD++~\cite{zhang2021beyond}, Zero-DCE~\cite{guo2020zero}, EnlightenGAN~\cite{jiang2021enlightengan}, and Self-Supervised~\cite{zhang2020self}. We have fine-tuned all models on the LOL train set and then evaluated it on the LOL test set. Fig.\ref{fig:LOL} shows some representative results for visual comparison. The enhanced results show that the EnlightenGAN and Zero-DCE fail to recover the images. HE significantly improves the brightness of the low-light image. However, it applies a contrast pull-up to each channel of RGB separately, which leads to color distortion (for example, the wall in Fig.\ref{fig:LOL}(b)). LIME enhances the images by directly estimating the illumination map, but this approach enhances both details and noise. Retinex-Net notably improves the visual quality of the low-light images, but it over-smoothes details, amplifies noise, and even produces color bias. It seems that the results of Self-Supervised, KinD++, and ours have better visual quality among all the methods. To further investigate the differences between these three methods, we have zoomed in the details inside the red and green bounding boxes. We can find from Fig.\ref{fig:LOL}(h) that Self-Supervised produces blurred results for the rotation switch in the red rectangle, while the results of KinD++ and ours show a better reconstruction. For the platform area in the green rectangle, we can see that the image estimated by Self-Supervised is corrupted, while KinD++ and our results are clearer. In summary, the best visual quality can be obtained with our proposed method and KinD++. Considering that KinD++ is a supervised method, this shows that our proposed unsupervised method is very effective.
\begin{table}[t]
\centering
\caption{Quantitative comparisons on the LOL test set in terms of PSNR, SSIM, and NIQE. The best result is in red, whereas the second-best results are in blue, respectively. T, SL, and UL represent the traditional method, supervised learning method, and unsupervised learning method, respectively.}
\begin{tabular}{c|c|c|c|c}
\hline
\textbf{Learning} &\textbf{Method} &\textbf{PSNR}$\uparrow$ &\textbf{SSIM}$\uparrow$ &\textbf{NIQE}$\downarrow$\\ \hline
 &Input &7.77 &0.191 &6.749\\ \hline
\multirow{2}{*}{T}
 &HE~\cite{pizer1990contrast} &14.95 &0.409 &8.427\\ 
 &LIME~\cite{guo2016lime} &17.18 &0.484 &8.221\\ \hline
\multirow{2}{*}{SL}
 &Retinex-Net~\cite{Chen2018Retinex} &16.77 &0.425 &8.879\\ 
 &KinD++~\cite{zhang2021beyond} &{\textcolor{red}{21.32}} &{\textcolor{red}{0.829}} &5.120\\ \hline
\multirow{4}{*}{UL}
 &Zero-DCE~\cite{guo2020zero} &14.86 &0.562 &7.767\\ 
 &EnlightenGAN~\cite{jiang2021enlightengan} &17.48 &0.652 &{\textcolor{blue}{4.684}}\\ 
 &Self-Supervised~\cite{zhang2020self} &19.13 &0.651 &4.702\\ 
 &Ours &{\textcolor{blue}{20.23}} &{\textcolor{blue}{0.790}} &{\textcolor{red}{3.780}}\\ \hline
\end{tabular}
\label{table:LOL}
\end{table}

\subsection{Quantitative Evaluation}
We have also quantitatively compared our method to the other state-of-the-art methods. We have fine-tuned all models on the LOL train set and then evaluated it on the LOL test set. As shown in Table~\ref{table:LOL}, the proposed method achieves the best performance with an average PSNR score of 20.23 dB, SSIM score of 0.79, and NIEQ score of 3.78 in unsupervised methods, which exceed the second-best unsupervised method (Self-Supervised) by 1.1 dB on PSNR, 0.139 on SSIM, and 0.922 on NIQE. It demonstrates that the proposed method possesses the highest capability among all unsupervised methods and its performance is approximating the level of the state-of-the-art supervised methods. Recently, NIQE has been used to evaluate the image quality of low-light image enhancement, which evaluating real image restoration without ground truth. A smaller NIQE score indicates better visual quality. We can see from Table~\ref{table:LOL} that our method obtains the best NIQE scores in all unsupervised methods and even surpasses the state-of-the-art supervised method KinD++. It indicates that the low-light images enhanced with our method have the best visual quality.
\begin{table*}[t]
\centering
\caption{NIQE scores on low-light image sets(MEF, LIME, NPE, VV, DICM, SCIE, ExDark, EnlightenGAN, COCO). The best result is in red whereas the second best results are in blue, respectively. Smaller NIQE scores indicate a better quality of perceptual tendency.}
\begin{tabular}{c|c|c|c|c|c|c|c|c|c|c|c}
\hline
\textbf{Learning} &\textbf{Method} &MEF &LIME &NPE &VV &DICM &EnlightenGAN &SCIE &ExDark &COCO &Avg\\ \hline
\multirow{2}{*}{T}
&HE~\cite{pizer1990contrast} &3.472 &4.125 &4.289 &3.202 &3.643 &6.993 &3.373 &4.135 &4.206 &4.530\\ 
&LIME~\cite{guo2016lime} &3.56 &4.138 &4.194 &2.456 &3.818 &6.956 &3.222 &4.759 &4.24 &4.667\\ \hline
\multirow{2}{*}{SL}
&Retinex-Net~\cite{Chen2018Retinex} &4.386 &4.68 &4.567 &2.461 &4.451 &8.063 &3.705 &5.274 &4.89 &5.296\\ 
&KinD++~\cite{zhang2021beyond} &3.734 &4.81 &4.381 &2.352 &3.787 &{\textcolor{blue}{4.572}} &3.143 &{\textcolor{blue}{4.074}} &{\textcolor{blue}{3.896}} &{\textcolor{blue}{3.926}}\\ \hline
\multirow{4}{*}{UL}
&Zero-DCE~\cite{guo2020zero} &3.283 &3.782 &4.273 &3.217 &3.56 &6.582 &3.284 &4.149 &3.903 &4.386\\ 
&EnlightenGAN~\cite{jiang2021enlightengan} &{\textcolor{blue}{3.221}} &{\textcolor{blue}{3.678}} &{\textcolor{blue}{4.125}} &{\textcolor{red}{2.251}} &{\textcolor{blue}{3.546}} &4.609 &{\textcolor{blue}{2.939}} &4.357 &3.953 &3.973\\
&Self-Supervised~\cite{zhang2020self} &4.477 &4.966 &4.743 &3.364 &4.588 &4.872 &3.978 &5.176 &4.947 &4.758\\ 
&Ours &{\textcolor{red}{3.188}} &{\textcolor{red}{3.484}} &{\textcolor{red}{3.504}} &{\textcolor{blue}{2.336}} &{\textcolor{red}{3.425}} &{\textcolor{red}{3.711}} &{\textcolor{red}{2.864}} &{\textcolor{red}{3.422}} &{\textcolor{red}{3.037}} &{\textcolor{red}{3.325}}\\ \hline
\end{tabular}
\label{table:real}
\end{table*}
\subsection{Generalization Ability on Real-World Images}
To further demonstrate the generalization ability of the proposed method, we have tested the proposed method on some real-world low-light image sets, including MEF~\cite{lee2011power}(17 images), LIME~\cite{guo2016lime}(10 images), NPE~\cite{wang2013naturalness}(84 images), VV\footnote{https://sites.google.com/site/vonikakis/datasets}(24 images), DICM~\cite{lee2013contrast}(64 images), EnlightenGAN~\cite{jiang2021enlightengan}(148 images), SCIE~\cite{cai2018learning}(select 100 low-light images from the datasets). Furthermore, in order to showcase this unique advantage of our method in practice, we also conduct experiments using low-light images from other datasets, which are built for object detection and recognition. We selected 216 low-light images from ExDark~\cite{loh2019getting} and 100 nighttime images from COCO~\cite{lin2014microsoft}. We have fine-tuned all models on the EnlightenGAN train set\footnote {Since EnlightenGAN dataset is unpaired and cannot be used as the train set for supervised method, so we use LOL dataset as the train set for the supervised method} and then evaluated it on all the low-light image sets. As all these datasets are unpaired, we employ the NIQE metric to provide quantitative comparisons with the state-of-the-art methods, which are used for evaluating real image restoration without ground truth. The NIQE results on nine publicly available image sets used by previous works are reported in Table~\ref{table:real}. Our method achieved the best performance in eight of these nine datasets and achieved the first place in the average score. Fig.\ref{fig:real} shows the results of a challenging image on the SCIE dataset. From the results, we can observe that our proposed method and KinD++ enhance dark regions and simultaneously preserve the color. The result is visually pleasing without obvious noise and color casts. In contrast, HE, LIME, and EnlightenGAN generate visually good results, but it contains some undesired artifacts (e.g., the white wall). Zero-DCE fails to recover the image. Retinex-Net and Self-Supervised over-smooth the details, amplify noise, and even produce color deviation. Our proposed method and KinD++ enhance dark regions and preserve the color of the input image simultaneously. The result is visually pleasing without obvious noise and color casts. It demonstrates that our method has great generalization ability in real-world images with more naturalistic quality.


\section{Ablation Study}
\label{ablation}
To demonstrate the effectiveness of each component proposed in Section~\ref{method}, we conduct several ablation experiments. We primarily analyze the components in our Light Up Module (LUM), which are the core contribution and play critical roles in this work.

\subsection{Contribution of Light Up}

\subsubsection{Effect of Histogram Equalization Prior}
Since histogram equalization prior is the main contribution in our work, a comparative assessment of its validity has been carried out. We employ the histogram equalization enhanced image instead as the reference image. We have evaluated different loss functions with the histogram equalization enhanced image: L1 loss $\mathcal{L}_{L1}$, MSE loss $\mathcal{L}_{MSE}$, and SSIM loss $\mathcal{L}_{SSIM}$, and max information entropy loss $\mathcal{L}_{max}$~\cite{zhang2020self}. The formulas of these losses are as follows:
\begin{equation}
\mathcal{L}_{L1} = \parallel R - H(I)\parallel_1
\end{equation}
\begin{equation}
\mathcal{L}_{MSE} = \parallel R - H(I)\parallel_2^2
\end{equation}
\begin{equation}
\mathcal{L}_{SSIM} = 1-SSIM(R, H(I))
\end{equation}
\begin{equation}
\mathcal{L}_{max} = \parallel \mathop{max}\limits_{c\in{R,G,B}}(R^c) - H(\mathop{max}\limits_{c\in{R,G,B}} (I^c))\parallel_1
\end{equation}
\noindent where $H(\cdot)$ stands for histogram equalization operation, $R$ represents the relfectance map, $I$ denotes the input low-light image, $R^c$ represents the max channel of relfectance map, $I^c$ represents the max channel of input low-light image.

\begin{figure*}[t]
\centering
    \subfigure[Input]{
        \includegraphics[width=0.16\linewidth]{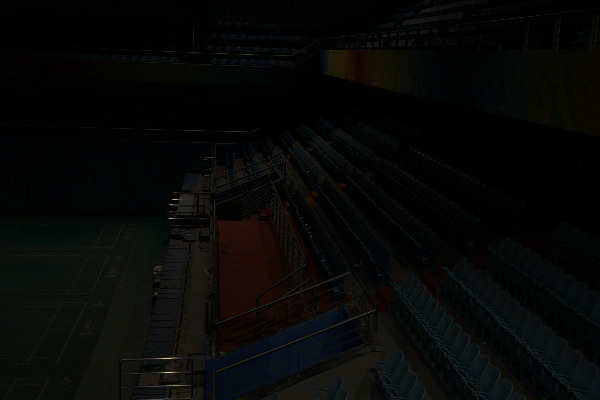}
    }\hspace*{-2mm}
    \subfigure[with $\mathcal{L}_{L1}$]{
        \includegraphics[width=0.16\linewidth]{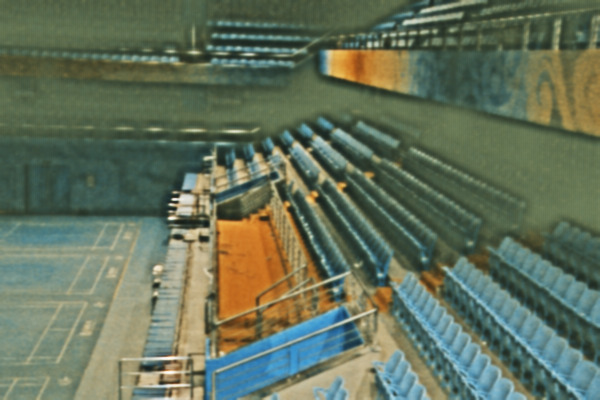}
    }\hspace*{-2mm}
    \subfigure[with $\mathcal{L}_{MSE}$]{
        \includegraphics[width=0.16\linewidth]{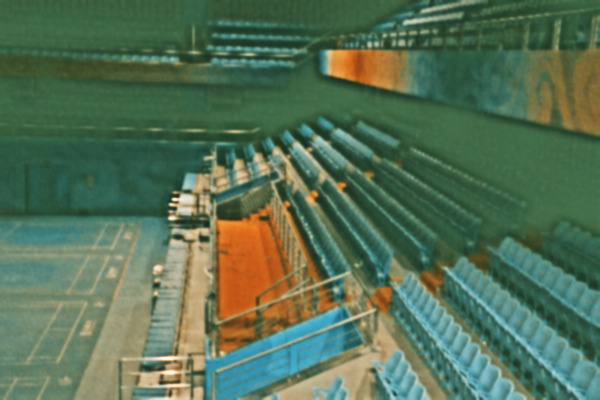}
    }\hspace*{-2mm}
    \subfigure[with $\mathcal{L}_{SSIM}$]{
        \includegraphics[width=0.16\linewidth]{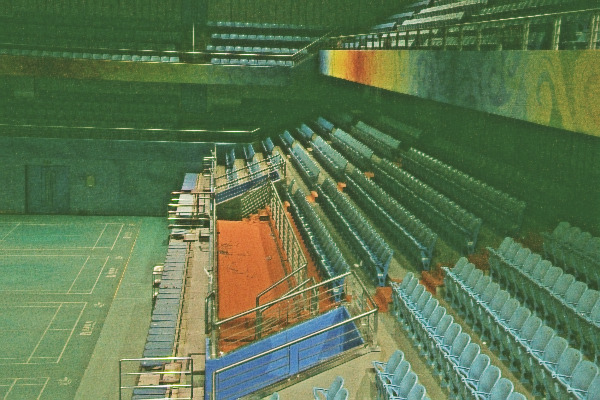}
    }\hspace*{-2mm}
    \subfigure[with $\mathcal{L}_{max}$]{
        \includegraphics[width=0.16\linewidth]{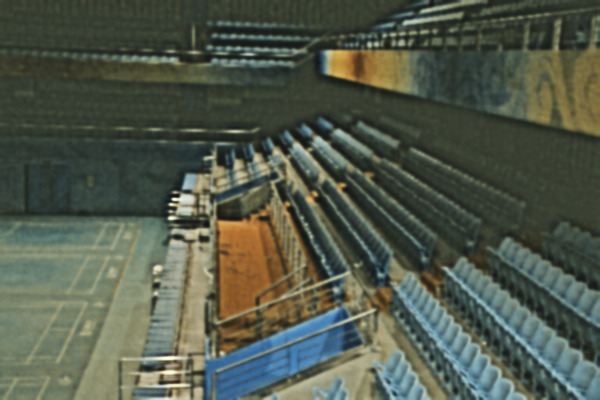}
    }\hspace*{-2mm}
    \subfigure[with HEP]{
        \includegraphics[width=0.16\linewidth]{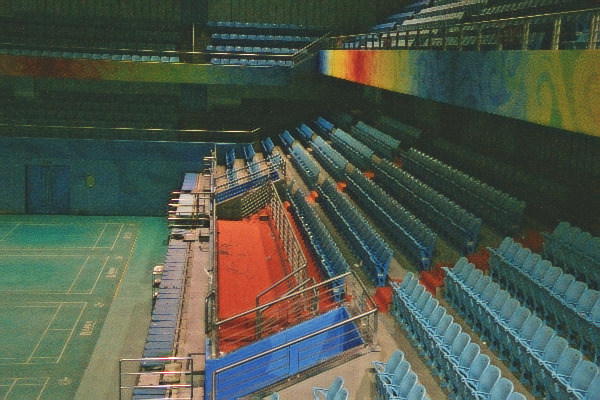}
    }
 \caption{Ablation study of the contribution of histogram equalization prior in LUM (replace reflectance similarity loss $\mathcal{L}_{rs}$ with L1 loss $\mathcal{L}_{L1}$, MSE loss $\mathcal{L}_{MSE}$, SSIM loss $\mathcal{L}_{SSIM}$, and max information entropy loss $\mathcal{L}_{max}$).}
 \label{Figure prior}
\end{figure*}

The comparison results are shown in Table \ref{table:prior}. Using $\mathcal{L}_{L1}$ or $\mathcal{L}_{MSE}$ achieves similar SSIM and NIQE scores. Nevertheless, for the PSNR values, the estimation from $\mathcal{L}_{MSE}$ exceeds those from $\mathcal{L}_{L1}$ with 0.33dB. The $\mathcal{L}_{SSIM}$ improves the NIQE score by a large margin. $\mathcal{L}_{SSIM}$ surpasses HEP in NIQE score, but HEP outperformed by 1.58dB in PSNR and 0.047 in SSIM. $\mathcal{L}_{max}$ achieves similar SSIM scores with HEP, but it failed in NIQE and PSNR by a large margin. Fig.\ref{Figure prior} shows a visual comparison of these loss functions. $\mathcal{L}_{L1}$ and $\mathcal{L}_{MSE}$ significantly improves the brightness of the low-light images. However, they have obvious color deviations (e.g., the color of the floor) and undesired artifacts (e.g., the dark region of the wall). $\mathcal{L}_{SSIM}$ reveals the color and texture, but with the blurry mask. $\mathcal{L}_{max}$ has color distortion. Both quantitative and qualitative results demonstrate the effectiveness of the proposed prior.

\begin{figure*}[htbp]
\centering
    \subfigure[Input]{
        \includegraphics[width=0.19\linewidth]{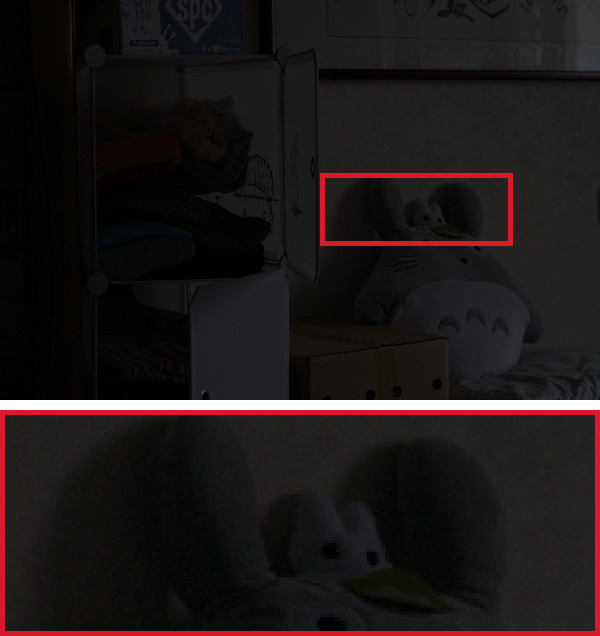}
    }\hspace*{-2mm}
    \subfigure[w/o $\mathcal{L}_{recon}$]{
        \includegraphics[width=0.19\linewidth]{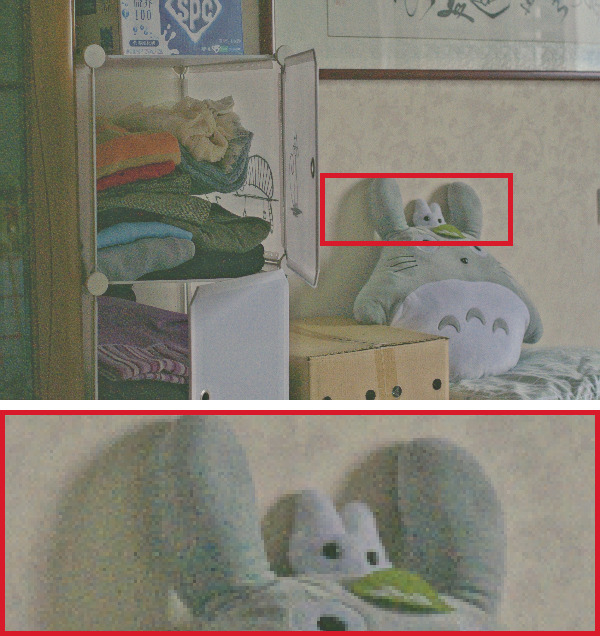}
    }\hspace*{-2mm}
    \subfigure[w/o $\mathcal{L}_{is}$]{
        \includegraphics[width=0.19\linewidth]{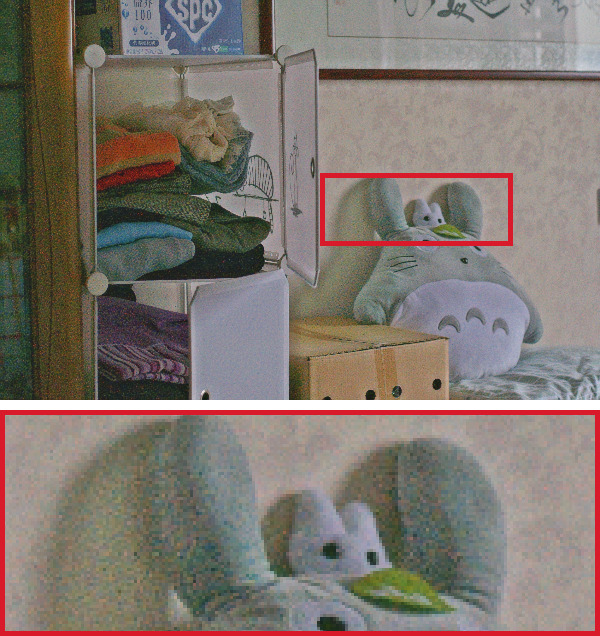}
    }\hspace*{-2mm}
    \subfigure[w/o $\mathcal{L}_{hep}$]{
        \includegraphics[width=0.19\linewidth]{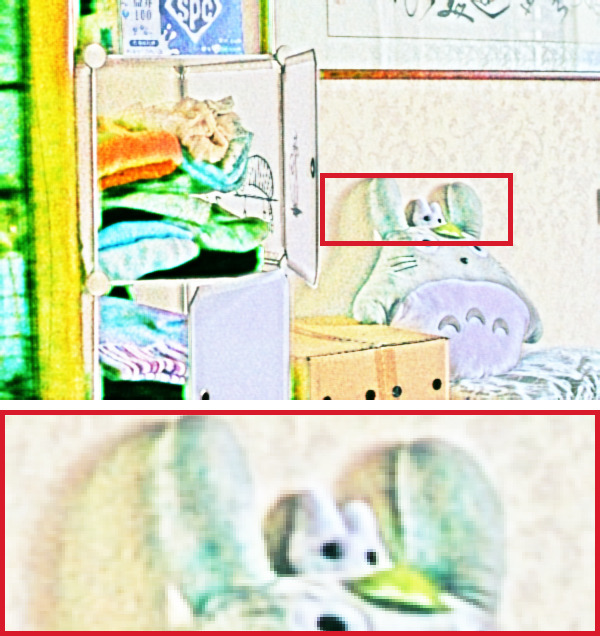}
    }\hspace*{-2mm}
    \subfigure[full loss]{
        \includegraphics[width=0.19\linewidth]{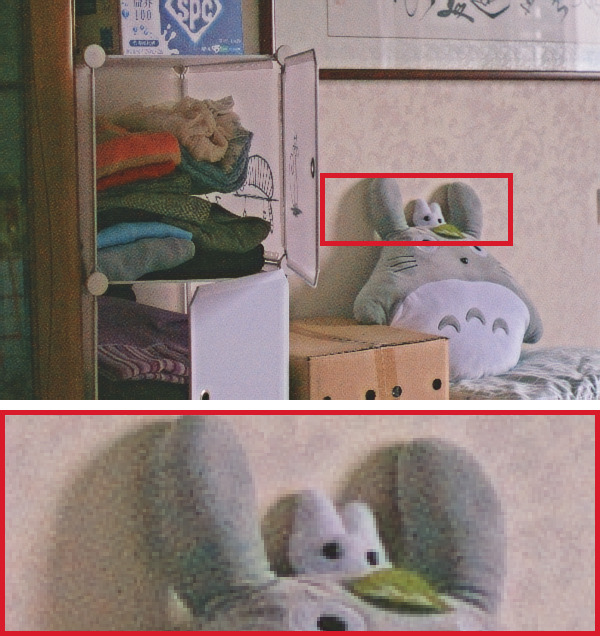}
    }
 \caption{Ablation study of the contribution of loss functions in LUM (reconstruction loss $\mathcal{L}_{recon}$, illumination smoothness loss $\mathcal{L}_{is}$), histogram equalization prior loss $\mathcal{L}_{hep}$). Red rectangle indicate the obvious differences and amplified details.}
 \label{Figure abs1loss}
\end{figure*}

\begin{table}[t]
\centering
\caption{Ablation study of the contribution of histogram equalization prior in LUM in terms of PSNR, SSIM and NIQE.}
\begin{tabular}{c|c|c|c}
\hline
\textbf{Loss Function} & \textbf{PSNR}$\uparrow$ & \textbf{SSIM}$\uparrow$ & \textbf{NIQE}$\downarrow$ \\ \hline
Input &7.77 &0.191 &6.749\\ 
with $\mathcal{L}_{L1}$ &17.51 &0.687 &6.343\\ 
with $\mathcal{L}_{MSE}$ &17.84 &0.698 &6.649\\ 
with $\mathcal{L}_{SSIM}$ &17.94 &0.654 &4.869\\ 
with $\mathcal{L}_{max}$ &18.29 &0.690 &7.294\\ 
with HEP &19.52 &0.701 &5.480\\ \hline
\end{tabular}
\label{table:prior}
\end{table}

\subsubsection{Effect of Loss functions}
We present the results of LUM trained by various combinations of losses in Fig.~\ref{Figure abs1loss}. Removing the reconstruction loss $\mathcal{L}_{recon}$ fails to brighten the image, and this shows the importance of reconstruction loss in enhancing the quality of the generated image. The results with illumination smoothness loss $\mathcal{L}_{is}$haves relatively lower contrast than the full results, and it shows smooth illumination map can somehow brighten the reflectance map. Finally, removing the histogram equalization prior loss $\mathcal{L}_{hep}$ hampers the correlations between neighboring regions leading to obvious artifacts. To further demonstrate the effectiveness of each loss, we conduct several experiments on the LOL dataset. The evaluation results of each loss show in Table~\ref{table:loss1}. The results show that without the histogram equalization prior loss, the PSNR decrease from 19.52 to 9.0, the SSIM decrease from 0.701 to 0.54. It demonstrates the importance of perceptual loss. To better prove the role of perceptual loss, we conduct an ablation study on this prior in the next subsection.
\begin{table}[t]
\centering
\caption{Ablation study of the contribution of loss functions in LUM in terms of PSNR, SSIM and NIQE.}
\begin{tabular}{c|c|c|c}
\hline
\textbf{Loss Function} &\textbf{PSNR}$\uparrow$ &\textbf{SSIM}$\uparrow$ &\textbf{NIQE}$\downarrow$\\ \hline
Input &7.77 &0.191 &6.749\\ 
w/o $\mathcal{L}_{hep}$ &9.00 &0.540 &4.539\\ 
w/o $\mathcal{L}_{recon}$ &17.06 &0.675 &6.782\\ 
w/o $\mathcal{L}_{is}$ &17.93 &0.621 &6.350\\
full loss &19.52 &0.701 &5.480\\ \hline
\end{tabular}
\label{table:loss1}
\end{table}

\subsection{Contribution of Noise Disentanglement}
\subsubsection{Effect of Network Architecture}
\begin{figure*}[htbp]
\centering
    \subfigure[Input]{
        \includegraphics[width=0.19\linewidth]{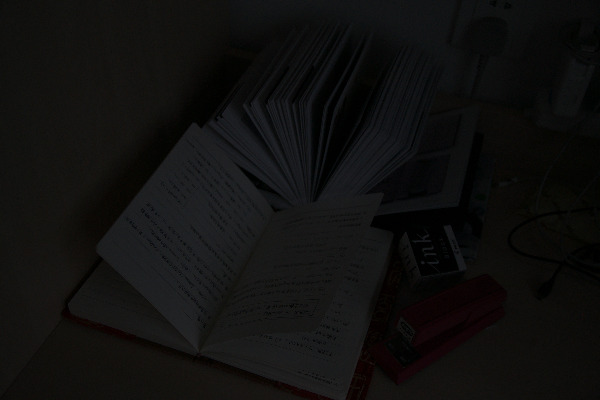}
    }\hspace*{-2mm}
    \subfigure[LUM + BM3D~\cite{dabov2007image}]{
        \includegraphics[width=0.19\linewidth]{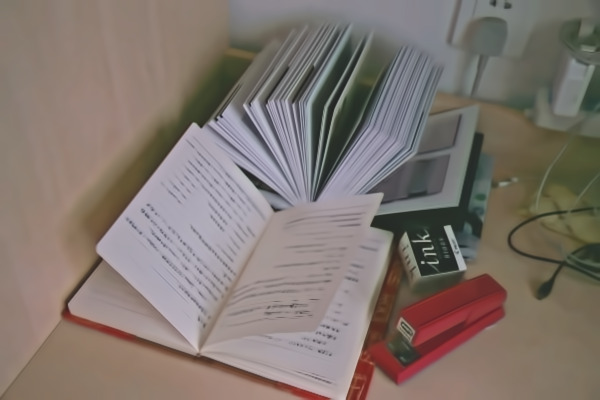}
    }\hspace*{-2mm}
        \subfigure[LUM + Du~\etal~\cite{du2020learning}]{
        \includegraphics[width=0.19\linewidth]{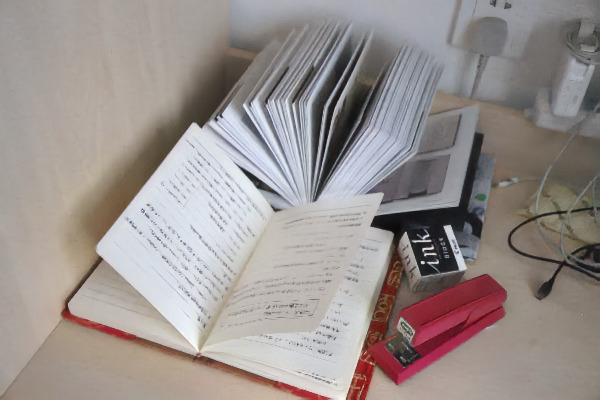}
    }\hspace*{-2mm}
    \subfigure[LUM + NDM]{
        \includegraphics[width=0.19\linewidth]{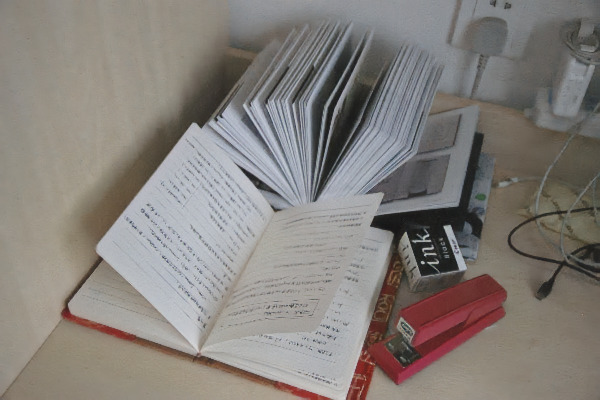}
    }\hspace*{-2mm}
    \subfigure[Ground-Truth]{
        \includegraphics[width=0.19\linewidth]{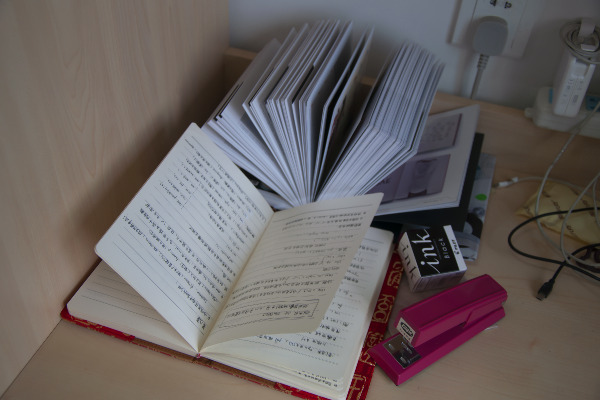}
    }
\caption{Ablation study of the contribution of noise encoder in NDM (compare with BM3D and a GAN-based denoise mode). Red rectangle indicate the obvious differences and amplified details.}
 \label{fig noise}
\end{figure*}

In this part, we compare three different denoise manners, including a traditional denoising tool BM3D~\cite{dabov2007image}, a GAN-based denoise method~\cite{du2020learning}, which has a similar architecture to ours, and our proposed NDM. Fig.\ref{fig noise} shows the comparison results of these three methods. The BM3D and the GAN-based method are the state-of-the-art denoising methods. However, the results show that the BM3D can handle noise, but it blurs the image. The GAN-based method is visually similar to our proposed NDM, but the image is overexposed compared to the ground truth. The result of our proposed NDM contains more delicate details and more vivid colors than other methods. As the quantitative results are shown in Table~\ref{table:arc2}, the NDM improves the GAN-based denoise method by a large margin in terms of PSNR and outperforms the BM3D by about 0.66dB in PSNR, 0.014 in SSIM, and 2.497 in NIQE. The new design of the NDM proves its effectiveness by the best results in this comparison.
\begin{table}[t]
\centering
\caption{Ablation study of the contribution of noise encoder in NDM in terms of PSNR, SSIM, and NIQE.}
\begin{tabular}{c|c|c|c}
\hline
\textbf{Denoise Model} &\textbf{PSNR}$\uparrow$ &\textbf{SSIM}$\uparrow$ &\textbf{NIQE}$\downarrow$ \\ \hline
LUM &19.52 &0.701 &5.480\\ 
LUM + BM3D~\cite{dabov2007image} &19.57 &0.776 &6.277\\ 
LUM + Du~\etal~\cite{du2020learning} &18.74 &0.791 &4.539\\ 
LUM + NDM &20.23 &0.790 &3.780\\ \hline
\end{tabular}
\label{table:arc2}
\end{table}
\subsubsection{Effect of Loss functions}
We evaluate the loss functions of the NDM and the evaluation results are shown in table~\ref{table:loss2}. From the results, we can conclude that removing self-reconstruction loss $\mathcal{L}_{recon}$ can significantly reduce PSNR and NIQE scores. Without the KL divergence loss $\mathcal{L}_{KL}$, background consistency loss $\mathcal{L}_{bc}$, and perceptual loss $\mathcal{L}_{per}$, all metrics have dropped a lot. Removing the adversarial loss $\mathcal{L}_{adv}$ cause SSIM and NIQE scores to drop a lot. Finally, when removing the cycle-consistency loss $\mathcal{L}_{cc}$, the NIQE scores have risen by 0.028, but at the same time, PSNR and SSIM have dropped by 0.32dB and 0.01. The entire loss function of NDM is designed to transfer noise image to clean image, and it performs stronger noise suppression on regions where the brightness is significantly promoted after the image brightness enhancement guided by the histogram equalization prior.
\begin{table}[t]
\centering
\caption{Ablation study of the contribution of loss functions in NDM in terms of PSNR, SSIM, and NIQE.}
\begin{tabular}{c|c|c|c}
\hline
\textbf{Loss Function} &\textbf{PSNR}$\uparrow$ &\textbf{SSIM}$\uparrow$ &\textbf{NIQE}$\downarrow$ \\ \hline
w/o $\mathcal{L}_{adv}$ &19.66 &0.705 &5.299\\ 
w/o $\mathcal{L}_{KL}$ &19.68 &0.778 &4.394\\ 
w/o $\mathcal{L}_{per}$ &19.83 &0.781 &4.389\\
w/o $\mathcal{L}_{cc}$ &19.91 &0.780 &3.752\\
w/o $\mathcal{L}_{bc}$ &19.92 &0.785 &4.143\\
w/o $\mathcal{L}_{recon}$ &19.96 &0.783 &4.234\\ 
full loss &20.23 &0.790 &3.780\\ \hline
\end{tabular}
\label{table:loss2}
\end{table}

\section{Conclusion}
\label{conclusion}
In this work, we propose an unsupervised network for low-light image enhancement. Inspired by Retinex theory, we design a two-stage network to enhance the low-light image. The first stage is an image decomposition network termed light up module (LUM), and the second stage is an image denoising network termed noise disentanglement module (NDM). The LUM brightens the image by decomposing the images into reflectance and illumination maps. In the absence of ground truth, we introduce an effective prior termed histogram equalization prior to guiding the training process, which is an extension of histogram equalization that investigates the spatial correlation between feature maps. Benefiting from the abundant information of the histogram equalization prior, the reflectance maps generated by LUM simultaneously improve brightness and preserve texture and color information. The NDM further denoises the reflectance maps to obtain the final images while preserving more natural color and texture details. Both qualitative and quantitative experiments demonstrate the advantages of our model over state-of-the-art algorithms.

In the future work, we intend to explore more effective prior for low-light image enhancement and investigate some GAN-based methods for low-light and normal-light image transfer. Due to the limited application value of low light enhancement, we also expect to integrate enhancement algorithms with some high-level tasks, such as object detection and semantic segmentation, which can be used for autonomous driving to provide reliable visual aids for dark and challenging environments.


\ifCLASSOPTIONcaptionsoff
  \newpage
\fi

\bibliographystyle{IEEEtran}
\bibliography{reference}

\begin{IEEEbiography}
[{\includegraphics[width=1in,height=1.25in,clip,keepaspectratio]{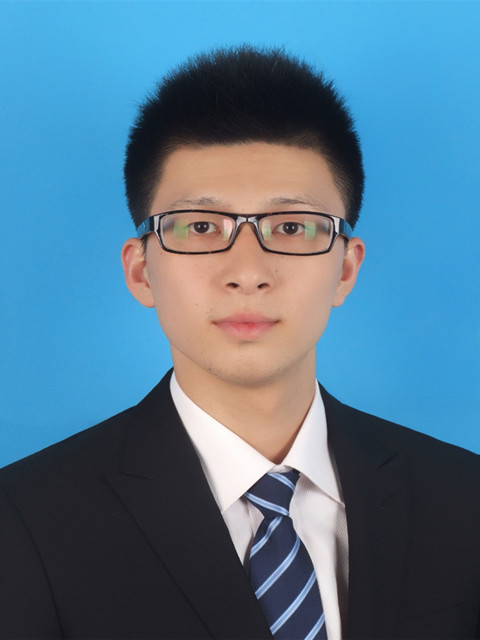}}]{Feng Zhang}
is currently a Ph.D. candidate in the School of Artificial Intelligence and Automation, Huazhong University of Science and Technology, supervised by Prof. Nong Sang. He received his M.S. degree from the School of Materials Science and Engineering, Huazhong University of Science and Technology in 2015. His research interests include computer vision and deep learning. Now he mainly works on the area of low-light image enhancement.
\end{IEEEbiography}

\begin{IEEEbiography}
[{\includegraphics[width=1in,height=1.25in,clip,keepaspectratio]{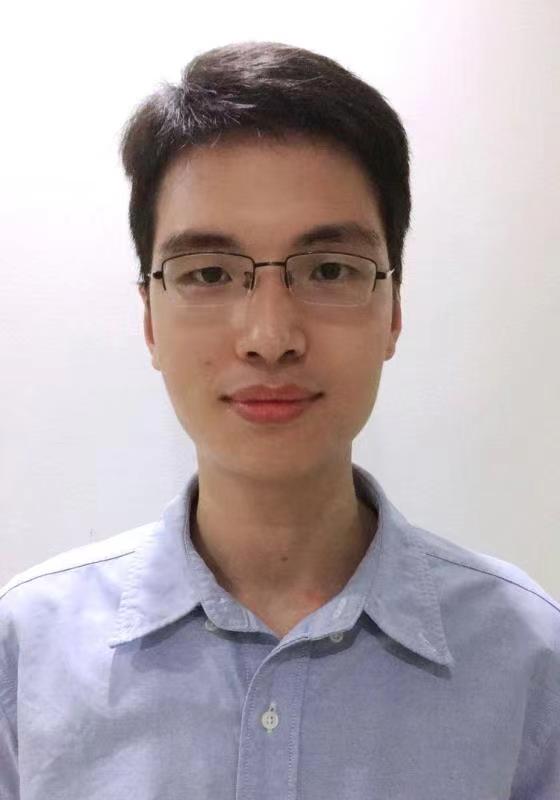}}]{Yuanjie Shao}
received the B.S. and M.S degree in college of mechanical and electronic information, China University of Geosciences in 2010 and 2013, Wuhan, China, and PhD degree in Control science and Engineering from Huazhong University of Science and Technology in 2018. He is currently a postdoctoral with the School of Artificial Intelligence and Automation, Huazhong University of Science and Technology, Wuhan, China. His research interests include pattern recognition, computer vision.
\end{IEEEbiography}

\begin{IEEEbiography}
[{\includegraphics[width=1in,height=1.25in,clip,keepaspectratio]{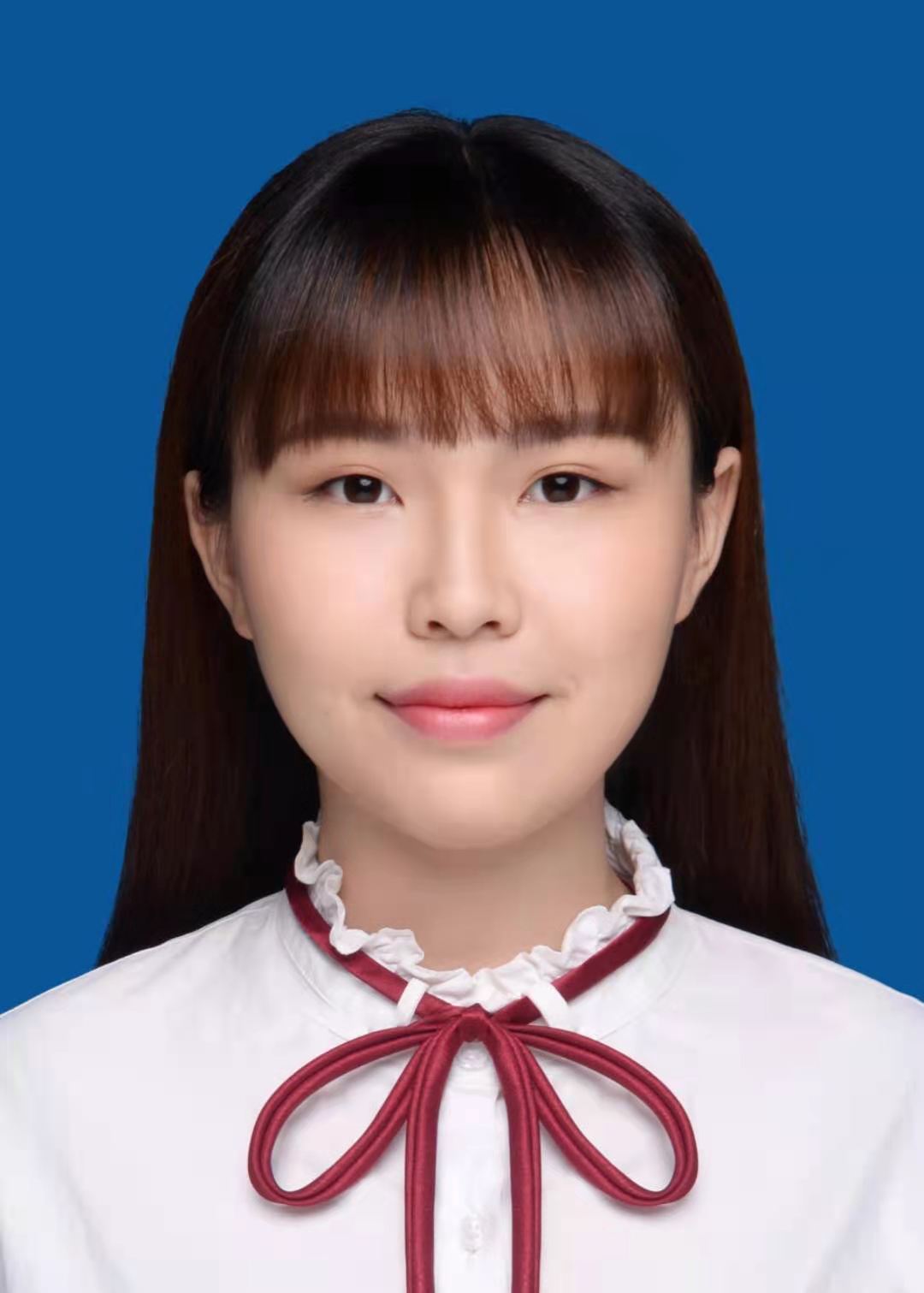}}]{Yishi Sun}
received the bachelor’s degree from the Northwestern Polytechnical University in 2020. She is currently pursuing the master degree with the Huazhong University of Science and Technology. Her research interests include computer vision and deep learning. Her main research field is image enhancement.
\end{IEEEbiography}

\begin{IEEEbiography}
[{\includegraphics[width=1in,height=1.25in,clip,keepaspectratio]{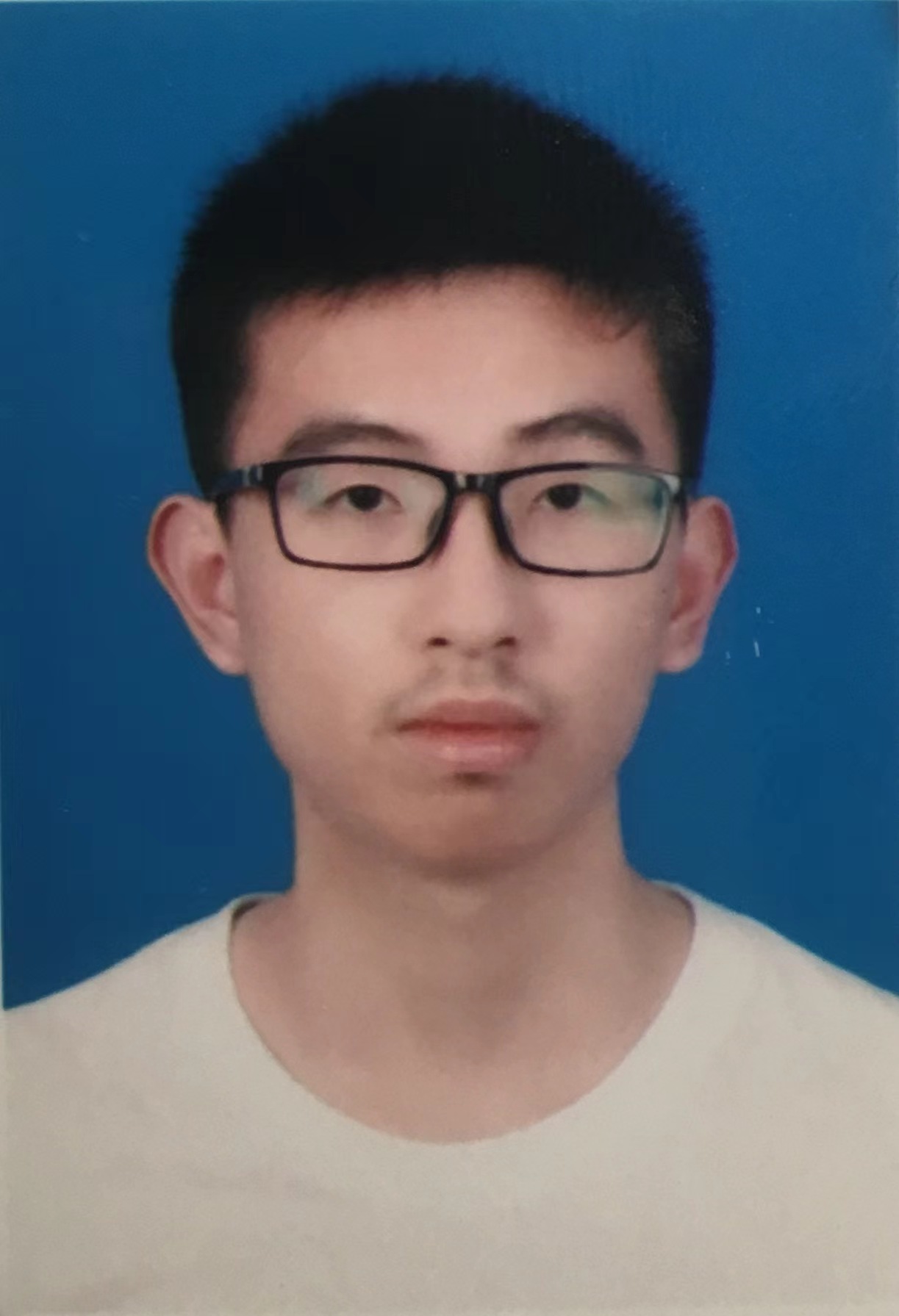}}]{Kai Zhu}
received the bachelor’s degree from the Huazhong University of Science and Technology in 2020. He is currently pursuing the master degree with the Huazhong University of Science and Technolog. His research interests include computer vision and deep learning. He mainly works on image and video deblurring.
\end{IEEEbiography}

\begin{IEEEbiography}
[{\includegraphics[width=1in,height=1.25in,clip,keepaspectratio]{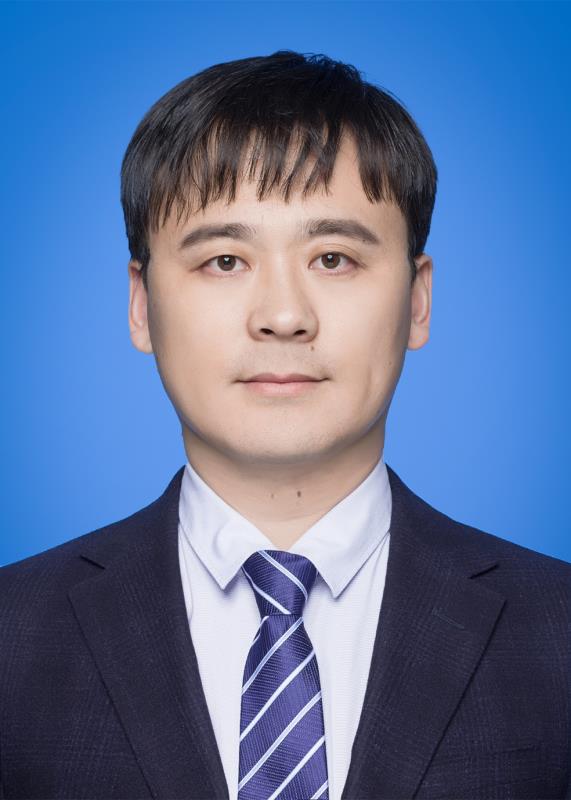}}]{Changxin Gao}
received the Ph.D. degree in pattern recognition and intelligent systems from the Huazhong University of Science and Technology in 2010. He is currently an Associate Professor with the School of Artificial Intelligence and Automation, Huazhong University of Science and Technology. His research interests are pattern recognition and surveillance video analysis.
\end{IEEEbiography}

\begin{IEEEbiography}
[{\includegraphics[width=1in,height=1.25in,clip,keepaspectratio]{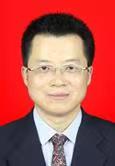}}]{Nong Sang}
received the Ph.D. degree in pattern recognition and intelligent control from the Huazhong University of Science and Technology in 2000. He is currently a Professor with the School of Artificial Intelligence and Automation, Huazhong University of Science and Technology. His research interests include pattern recognition, computer vision, and neural networks.
\end{IEEEbiography}
\end{document}